\documentclass[10pt,journal,compsoc]{IEEEtran}
\ifCLASSOPTIONcompsoc
  \usepackage[nocompress]{cite}
\else
  \usepackage{cite}
\fi
\ifCLASSINFOpdf
  \usepackage[pdftex]{graphicx}
\else
\fi
\usepackage{amsmath}
\usepackage{amssymb}
\usepackage{times}
\usepackage{epsfig}
\usepackage{multirow}
\usepackage{color,soul}
\usepackage{enumitem}
\usepackage{microtype}
\usepackage{mathabx}
\usepackage{stfloats}

\begin{document}
%
\title{Contextual Translation Embedding for Visual Relationship Detection and Scene Graph Generation}
%
%
%

\author{Zih-Siou~Hung, 
        Arun~Mallya, 
        and~Svetlana~Lazebnik 
\IEEEcompsocitemizethanks{\IEEEcompsocthanksitem Zih-Siou Hung and Svetlana Lazebnik  are with the Computer Science Department, University of Illinois at Urbana-Champaign, Urbana, IL 61801. Email: zhung2@illinois.edu and slazebni@illinois.edu
\IEEEcompsocthanksitem Arun Mallya is with Nvidia Research. Email: amallya2@illinois.edu }
}

\IEEEtitleabstractindextext{%
\begin{abstract}
Relations amongst entities play a central role in image understanding. Due to the complexity of modeling (\emph{subject}, \emph{predicate}, \emph{object}) relation triplets, it is crucial to develop a method that can not only recognize seen relations, but also generalize 
to unseen cases. Inspired by a previously proposed visual translation embedding model, or
VTransE~\cite{Zhang_2017_CVPR}, we propose a context-augmented translation embedding model that 
can capture both common and rare relations. 
The previous 
VTransE model maps entities and predicates into a low-dimensional embedding vector 
space where the predicate is interpreted as a translation vector between the embedded 
features of the bounding box regions of the \emph{subject} and the \emph{object}. Our model additionally incorporates the contextual information captured by the bounding box of the \emph{union} of the subject and the object, and learns the embeddings guided by the constraint \emph{predicate} $\approx$ \emph{union} (\emph{subject}, \emph{object}) $-$ \emph{subject} $-$ \emph{object}.
In a comprehensive evaluation on multiple challenging benchmarks, our approach outperforms previous translation-based models and comes close to or exceeds the state of the art across a range of settings, from small-scale to large-scale datasets, from common to previously unseen relations. It also achieves promising results for the recently introduced task of scene graph generation.
\end{abstract}

}

\maketitle

\IEEEdisplaynontitleabstractindextext

%
\IEEEpeerreviewmaketitle

\IEEEraisesectionheading{\section{Introduction}\label{sec:intro}}

Performance on object detection and localization has improved greatly over the last few years with the introduction of the deep R-CNN model~\cite{girshick2014rich} and its successors ~\cite{girshick15fastrcnn,he2017mask,ren2015faster,redmon2017yolo9000}. The next natural step is to go beyond detecting individual objects and start reasoning about semantic relationships between multiple objects, which could be useful for applications such as image captioning~\cite{lu2018neural}, retrieval~\cite{sg2015,PrabhuICCV2015}, and visual question answering~\cite{nmn2016}.

In this work, we address the task of Visual Relationship Detection (VRD)~\cite{lu2016visual}, which focuses on understanding interactions between pairs of object entities in the image. 
These interactions can be spatial, comparative, or action-based, and are represented as (\emph{subject}, \emph{predicate}, \emph{object}) triplets such as (\emph{desk}, \emph{beneath}, \emph{laptop}), (\emph{tower}, \emph{taller than}, \emph{trees}), or (\emph{person}, \emph{eat}, \emph{pizza}). VRD has two goals: detection of object instances participating in an interaction, and correct prediction of the interaction type.
Inferring the relations between object pairs is not always straightforward visually, and depends on context. For instance, (\emph{person}, \emph{hold}, \emph{umbrella}) and (\emph{person}, \emph{hold}, \emph{guitar}) are dissimilar in an image even though they share the same predicate `\emph{hold}'.
The very large output space makes this task even more challenging. 
Consider the Stanford VRD dataset~\cite{lu2016visual}, which has $100$ classes of objects, $70$ classes of predicates, and a total of 30k training relationship annotations. The number of possible interaction triplets, including unusual cases such as (\emph{dog}, \emph{ride}, \emph{horse}), is $100 \times 100 \times 70 = 700k$, meaning that most relationships do not even have a training example. This sparsity necessitates the development of methods that can recognize the predicate even if it occurs with a novel subject or object. 

To improve generalization to rare or unseen relationships, we propose a novel framework called Union Visual Translation Embedding, or {\bf UVTransE}. Our starting point is the recently introduced {\bf VTransE} method of Zhang et al.~\cite{Zhang_2017_CVPR}, which maps entities and predicates into a low-dimensional embedding vector space where the \emph{predicate} is interpreted as a translation vector between the embedded appearance features of the \emph{subject} and the \emph{object}. 
More concretely, if $\boldsymbol{s}$, $\boldsymbol{p}$, and  $\boldsymbol{o}$ are vectors representing the subject, the predicate, and the object in the learned embedding space, VTransE assumes that a relationship ($\boldsymbol{s}$, $\boldsymbol{p}$, $\boldsymbol{o}$) exists if $\boldsymbol{s} + \boldsymbol{p} \approx \boldsymbol{o}$. This formulation was inspired, in turn, by translation embeddings for relational data~\cite{bordes2013}.

VTransE does a good job of predicting relationships that it has seen in training time; however, it is not well-suited to recognizing unseen relationship triplets. This is due to two critical issues. First, VTransE calculates object vectors based on the features from subject and predicate only. That is, subject and predicate vectors ($\boldsymbol{s}$, $\boldsymbol{p}$) in the learned embedding space completely determine the object $\boldsymbol{o}$ as $\boldsymbol{s} + \boldsymbol{p}$. Consider an unusual relationship, such as (\emph{dog}, \emph{drive}, \emph{car}). Since the triplet is rare in the training set, the \emph{dog} and \emph{drive} vectors are not trained to produce this particular object \emph{car}, so we end up with $\boldsymbol{s} + \boldsymbol{p} \not \approx \boldsymbol{o}$. In addition, the VTransE embedding is not sufficiently flexible for modeling cases where many possible objects can satisfy a predicate with a given subject, since fixing $\boldsymbol{s}$ and $\boldsymbol{p}$ roughly determines $\boldsymbol{o}$.

\begin{figure*}[ht]
\centering
\includegraphics[width=0.8\textwidth,height=0.25\textheight]{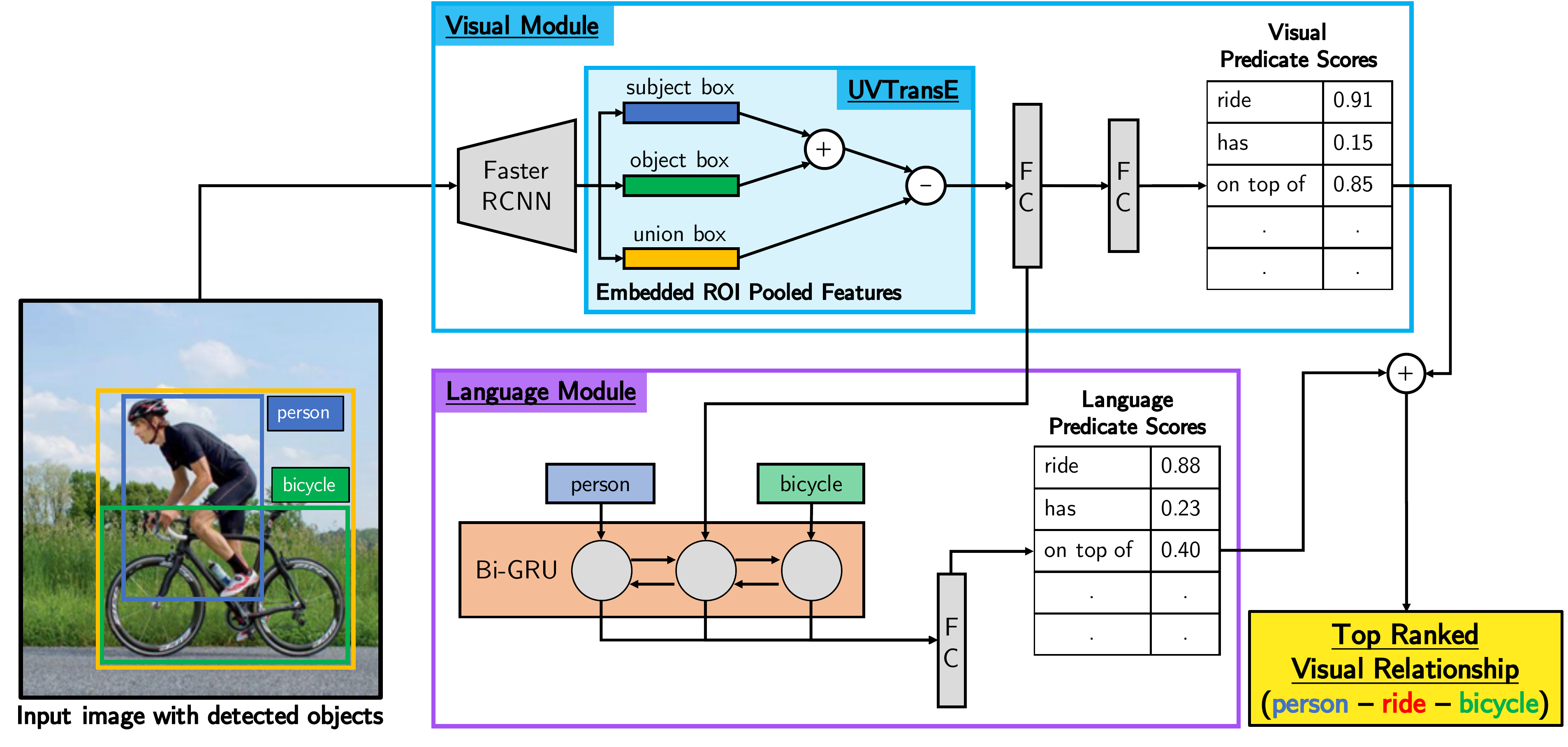}
\caption{Overview of our UVTransE visual relationship detection model. Given an image, Faster R-CNN is fist used to detect objects. For each pair of detected objects, appearance and spatial features are extracted and fed into the visual module, which computes the UVTransE embedding: \emph{union} $-$ (\emph{subject} + \emph{object}). The predicate embedding output by UVTransE may be optionally sent to a Bi-GRU language model. Finally, triplets ($s$, $p$, $o$) are ranked based on scores from the visual, language, and object detection modules.}
\label{fig:full_model}
\end{figure*}

In order to overcome the above two problems, we propose an extension to VTransE that not only enables triplets to be recognized in unseen cases, but also enables entities to have a distributed representation in the embedding space. Like VTransE, we model objects and predicates as embedding vectors; however, our predicate embedding vectors are not constrained to represent the translation between the subject and the object. Our idea is that by subtracting the embeddings of the \emph{subject} and the \emph{object} from the embedding of the entire box of the union of \emph{subject} and \emph{object} should provide an embedding corresponding to the predicate of interest, or $\boldsymbol{u} - \boldsymbol{s} - \boldsymbol{o} \approx \boldsymbol{p}$. For example, emb(\emph{person $\cup$ horse}) $-$ emb(\emph{person}) $-$ emb(\emph{horse}) $\approx$ emb(\emph{ride}). (Note that here and in the following, whenever we talk about the \emph{union box} or \emph{union feature}, we mean the bounding box of the union of the subject and object, and the features extracted from this box.) By removing object-related information from the contextual union feature, we hope to leave behind an embedding that contains information only about the predicate, leading to better zero-shot performance. Even though our modification of the VTransE formulation may seem straightforward, our experiments will demonstrate that UVTransE model can much better handle the challenges of VRD. For example, as shown in the results in Figure \ref{fig:vrd_res}, the learned predicate embedding `\emph{touch}' can model both (\emph{person}, \emph{touch}, \emph{skateboard}) and (\emph{person}, \emph{touch}, \emph{glasses}), even when (\emph{person}, \emph{touch}, \emph{glasses}) has not been seen during training. 

Similar to prior works like \cite{lu2016visual,plummerPLCLC2017}, we also incorporate a recurrent language model that uses word embeddings to learn about the semantic relatedness between different objects or different relations in an attempt to counteract the data sparsity problem. It has been shown that words with similar meaning are close to each other in word embedding spaces such as
word2vec~\cite{NIPS2013_5021} and GloVe~\cite{pennington2014glove}. Such semantic similarity might help us in detecting relation triplets not seen during training. For instance, given that we have seen {\em (person, ride, motorbike)} during training time, at test time, if we have an image containing the relation {\em (person, ride, bicycle)}, we might be able to detect this
relationship since motorbike and bicycle are semantically similar. Accordingly, we design a language module that benefits the overall detection task, including zero-shot cases. An overview of our UVTransE model, and its relationship with the language model, are shown in Fig.~\ref{fig:full_model}. Technical details will be given in Section~\ref{sec:uvtranse}.

In Section \ref{sec:exp} we will present an extensive empirical evaluation of our method on multiple datasets and settings, from small-scale to large-scale, and from common relationships to zero-shot recognition. In particular, we decisively outperform VTransE and most other competing methods on both the general and zero-shot settings of the VRD dataset~\cite{lu2016visual}, UnRel dataset~\cite{Peyre17}, two subsets of Visual Genome~\cite{Krishna:2017:VGC:3088990.3089101}, and the Open Images Challenge~\cite{OpenImages}. On the latter two datasets, we also apply our methods to the recently proposed task of {\em scene graph generation}.
A scene graph, introduced by Johnson \emph{et al.}~\cite{sg2015}, encapsulates all the relations amongst the object entities in an image. Its nodes correspond to objects and directed edges correspond to their pairwise relationships. 
We generate scene graphs via a simple two-stage approach, where we first detect objects, or nodes, and then infer relationships, or edges, using our UVTransE approach. Our experiments will show that this approach is competitive with more sophisticated state-of-the-art approaches designed to jointly reason about multiple edges of the graph, 
such as Neural Motifs~\cite{zellers2018scenegraphs}. As confirmed by all these experiments, our method is simple yet versatile and high-performing, which makes it a good choice for downstream applications that require prediction of relationships or scene graphs. Our code will be made publicly available.

\section{Related Work}

Detecting visual relationships in one form or another has been an active area of recognition research for at least the last decade. Most earlier works focused on predicting specialized types of predicates such as spatial relations  \cite{Galleguillos08, Gould2008}, or targeted human-centric relationships  \cite{gkioxari2015rstarcnn,Guadarrama:2013,mallya2016learning,maji2011action,yao2011human,YaoF10}. Such phrase or relationship detections were used in applications such as object recognition  \cite{choi_cvpr10, Papazoglou:2016:DOA:2997676.2998110,SalakhutdinovTT11}, image classification~\cite{MensinkGS14}, and text grounding \cite{fukui16emnlp,Plummer:2017:_flickr}. 

Recently, Lu \emph{et al.}~\cite{lu2016visual} introduced the generic visual relationship detection (VRD) task and a dataset that became one of the main benchmarks. They also proposed a VRD method that established the basic template for many follow-up works, including ours: first objects are detected, then object pairs are fed to a classifier that combines their appearance features with a language prior on the relationship triplet occurrence. Zhang \emph{et al.}~\cite{Zhang_2017_CVPR} projected features from the detected objects into a low-dimensional space and predicted the relationship using a learned relation translation vector. This VTransE method is the main departure point for our own work. Dai \emph{et al.}~\cite{dai2017detecting} proposed a deep relational network method exploiting the statistical dependencies between objects and their relationships, while Liang \emph{et al.}~\cite{vrl17} proposed building a semantic action graph capturing possible relations and learning to traverse it using a reinforcement learning formulation. 
In Zhang \emph{et al.}~\cite{zhang2018large}, the authors employed a novel triplet-softmax loss to learn the joint visual and semantic embedding.
Very recently, Zhang \emph{et al.}~\cite{zhang2019vrd} defined margin-based losses to address common types of errors existing in relationship prediction, resulting in a method that performs remarkably well on the detection of common relationship triplets, but does not necessarily generalize to rare or zero-shot relationships. 
As part of its visual representation, this method also uses the bounding box of the union of subject and object, however, it does not use a subtractive model for combining the union with the subject and object boxes, as we propose. 
Zhuang \emph{et al.}~\cite{zhuang2017towards} designed a context-aware interaction classifier with good generalization to the zero-shot case. Plummer \emph{et al.}~\cite{plummerPLCLC2017} obtained strong zero-shot performance through the use of multiple visual-language cues learned with Canonical Correlation Analysis (CCA). 
Yu \emph{et al.}~\cite{yu2017_vrd_knowledge_distillation} used a large amount of external textual data to distill useful knowledge for triplet learning. Peyre \emph{et al.}~\cite{Peyre17} focused on weakly supervised learning of relationships (not a setting we consider), and also introduced the UnRel dataset exhaustively annotated for a set of unusual triplets such as {\em (elephant, wear, glasses)}. This is one of the benchmarks used in our work.


As stated in the Introduction, we also apply our UVTransE method to scene graph generation. Most scene graph generation methods consider the surrounding context of a node as a valuable cue, and apply context propagation mechanisms to exchange information between neighboring nodes over a candidate scene graph. In Xu \emph{et al.}~\cite{xu2017scenegraph}, two sub-graphs, representing objects and relationships respectively, are created. Node features, which are used to predict relation types, are updated based on the messages passed between the two graphs. Similarly, Li \emph{et al.}~\cite{li2017msdn} proposed constructing a dynamic graph, where messages are passed across different feature representations to refine the scene graph. Zellers \emph{et al.}~\cite{zellers2018scenegraphs} designed a Stacked Motif Network to extract contextual cues, which are propagated across objects and relations. Yang \emph{et al.}~\cite{jwyang_graph_rcnn} developed an attentional graph convolutional network to place attention on reliable edges when information is exchanged between vertices in the candidate scene graph. In Sections \ref{subsec:vg} and \ref{subsec:openimages}, we apply our method to generate scene graphs on Visual Genome and Open Images datasets in a very straightforward way: we first run object detectors to find the nodes of the scene graph, and then use UVTransE to find the relations. Even though we are predicting each relationship independently, we will show that our results are competitive with those of more context-aware methods.

\section{The UVTransE Method}
\label{sec:uvtranse}

In our work, we split the VRD task into two stages. In the first stage, we use an off-the-shelf object detection model, such as Faster R-CNN~\cite{ren2015faster}, to predict object bounding boxes and per-class confidences in an image. For the second stage, we learn a model to score all possible triplets $(s,p,o)$ where $s$ is a {\em subject} box, $p$ is a {\em predicate} or relation label, and $o$ is an {\em object} box. Next, we describe our UVTransE relationship scoring model, which is illustrated in Figure~\ref{fig:full_model}.

\subsection{Union Visual Translation Embedding}
\label{subsec:uvtranse_def}
Let $\boldsymbol{s}$, $\boldsymbol{o}$, $\boldsymbol{u} \in \mathbb{R}^n $ be the appearance features of the bounding boxes enclosing the \emph{subject}, \emph{object}, and \emph{union} of subject and object, respectively. We want to learn three 
projection functions  $\boldsymbol{f}_s: \mathbb{R}^n \to \mathbb{R}^d$, $\boldsymbol{f}_o: \mathbb{R}^n \to \mathbb{R}^d$ and  $\boldsymbol{f}_u: \mathbb{R}^n \to \mathbb{R}^d$ 
that map the respective feature vectors into a common $d$-dimensional embedding space, as well as translation vectors $\boldsymbol{p} $ in the same space corresponding to each of the predicate labels present in the data. In our implementation, the functions $\boldsymbol{f}_s$, $\boldsymbol{f}_o$, and $\boldsymbol{f}_u$ are multilayer perceptrons. A relationship ($s$, $p$, $o$) that exists in the training data should impose the constraint
$\boldsymbol{f}_u (\boldsymbol{u}) - \boldsymbol{f}_s (\boldsymbol{s}) - \boldsymbol{f}_o (\boldsymbol{o}) \approx \boldsymbol{p}$.
To achieve this, similarly to~\cite{Zhang_2017_CVPR}, we learn 
$\boldsymbol{f}_s$, $\boldsymbol{f}_o$, $\boldsymbol{f}_u$, and
$\boldsymbol{p}$ 
by minimizing the following multi-class cross-entropy loss function:

\begin{equation}
    L_{\mathrm{vis}} = \sum_{\left(s, p, o\right) \in T} -\log  \frac{\mathrm{exp}\left(\boldsymbol{p}^\top \hat{\boldsymbol{p}}\right)}{\sum_{q \in P} \mathrm{exp}\left(\boldsymbol{q}^\top \hat{\boldsymbol{p}}\right)}, \label{eq:vis}
    \end{equation}
where
\begin{equation}
\hat{\boldsymbol{p}} =  \boldsymbol{f}_u (\boldsymbol{u}) - \boldsymbol{f}_s (\boldsymbol{s}) - \boldsymbol{f}_o (\boldsymbol{o})\,, \label{eq:hatp}
\end{equation}


\noindent $T$ is the set of all relationship triplets existing in the training data,\footnote{If there are multiple training examples with the same $(s,p,o)$, they yield multiple terms in the summations of Eqs. (1) and (3).} and $P$ is the set of all predicate labels. 
In practice, we found that we need to constrain the norms of 
$\boldsymbol{f}_u (\boldsymbol{u})$,
$\boldsymbol{f}_s (\boldsymbol{s})$,
and $\boldsymbol{f}_o (\boldsymbol{o})$
from getting arbitrary large. To this end, we augment Eq.~(\ref{eq:vis}) with soft constraints on embedding weights:
\begin{align}
L_{\mathrm{vis}} &= \sum_{\left(s, p, o\right) \in T}  -\log  \frac{\mathrm{exp}\left(\boldsymbol{p}^\top \hat{\boldsymbol{p}}\right)}{\sum_{q \in P} \mathrm{exp}\left(\boldsymbol{q}^\top \hat{\boldsymbol{p}}\right)} + \nonumber \\
& C \big( [\left\Vert 
\boldsymbol{f}_s (\boldsymbol{s})\right\Vert_2^2 - 1]_{+} + [\left\Vert 
\boldsymbol{f}_o (\boldsymbol{o})
\right\Vert_2^2 - 1]_{+} + \nonumber \\
& \quad\ [\left\Vert 
\boldsymbol{f}_u (\boldsymbol{u})
\right\Vert_2^2 - 1]_{+} \big), 
\label{eq:C}
\end{align}
where $[x]_{+} = \mathrm{max}(0, x)$. We experimented with other penalties to encourage the norms to stay close to one but found this one gave the best results. $C$ is a hyperparameter that determines the relative importance of the soft constraints, and its effect will be examined in Section~\ref{sec:exp}.

Our formulation of Eq. (\ref{eq:C}) differs from VTransE~\cite{Zhang_2017_CVPR} in the addition of the contextual union feature and the norm regularization terms. Ablation studies of Section \ref{subsec:vrd} will show that these modifications are key to improving performance, not only on common cases but also on the zero-shot case.

At test time, given a candidate triplet ($s$, $p$, $o$), we can score the predicate $p$ as
\begin{equation}
z_p = \frac{\mathrm{exp}\left(\boldsymbol{p}^\top \hat{\boldsymbol{p}}\right)}{\sum_{q \in P} \mathrm{exp}\left(\boldsymbol{q}^\top \hat{\boldsymbol{p}}\right)} \,.
\label{eq:z_p}
\end{equation}
Similarly to~\cite{Zhang_2017_CVPR}, we can then define the score of the entire triplet by the sum of softmax detection scores for the subject and object, $\left( z_s, z_o \right)$, and the above predicate score $z_p$:
\begin{equation}
z_{\left(s, p, o\right)} = z_s + z_p + z_o\,. 
\label{eq:vis_only_score}
\end{equation}
Alternatively, for some datasets, we obtained better performance by taking the product of the above scores. Dataset-specific details will be given in Section \ref{sec:exp}.

\subsection{Language Module} \label{subsec:language_module}
Similar to prior work~\cite{vrl17,lu2016visual,plummerPLCLC2017,yu2017_vrd_knowledge_distillation}, we combine UVTransE with a language model that helps to combat data sparsity and learns which relationships are plausible between pairs of object classes. 
Our language module is a bi-directional GRU (Bi-GRU)~\cite{Schuster:1997:BRN:2198065.2205129} that receives encodings of subject, predicate, and object in three successive steps, concatenates the hidden states, and uses them for predicate classification. Further details will be given in \ref{sec:overall_pipeline}.
The loss for our language module $L_{\mathrm{lang}}$ is a standard multi-class cross-entropy loss which encourages it to produce the ground truth predicate.
The combined loss for our model is given by
\begin{equation}
\label{eq:total}
L_{\mathrm{total}} = \alpha L_{\mathrm{vis}} + (1 - \alpha) L_{\mathrm{lang}} \,.
\end{equation}
The score $z_{\left(s, p, o\right)}$ for a candidate relationship is now given by
\begin{align}
z_{\left(s, p, o\right)} &= z_s + z_o + \alpha z_{p} + (1 - \alpha) z_{l}\, ,
\label{eq:alpha}
\end{align}
where $\alpha$ is the weight for the visual module and $z_{l}$ is the softmax predicate score from the language module. The values of $\alpha$ used in the experiments will be given in Section \ref{sec:exp}.

\subsection{Implementation Details}
\label{sec:overall_pipeline}
In detail, the stages of our pipeline are: object detection, extraction of appearance and location features from bounding boxes, UVTransE relation embedding, language module (optional), and relationship scoring. The implementation of each of these components is described below.

\smallskip
\noindent{\bf Object Detection.}
Our first step is to run an object detector to locate a set of candidate objects in an image. We train a separate Faster R-CNN detector~\cite{ren2015faster} for each dataset. Our experiments use two backbones: VGG-16~\cite{simonyan14VGG} and ResNet-101~\cite{he2016deep} (see Section \ref{sec:exp} for dataset-specific details).
Each candidate object output by the detector is associated with a bounding box $b_i$, object class probability $z_i$, and an ROI-pooled feature vector $f_i$.

\smallskip
\noindent{\bf Appearance feature extraction.}
Our appearance features are based on the 
ROI-pooled features $f_i$ obtained from the object detector. These are 4096-d for the VGG backbone and 2048-d for the ResNet backbone. More specifically, we use the ROIAlign features of~\cite{he2017mask}, although in our experience, the improvement they give over standard ROIPool features is slight (less than a percentage point in mAP and relationship detection measures). We follow the specification in section \ref{subsec:uvtranse_def}, and pass the features $f_i$ through two FC layers with ReLU activation. The output dimensionalities of FC layers are 512 and 256, and we get 256-d appearance features at the end.

\smallskip
\noindent{\bf Location feature extraction.}
We encode each single bounding box (subject or object) into a 5-d vector $l_i = (\frac{x_i}{W_I}, \frac{y_i}{H_I}, \frac{x_i + w_i}{W_I}, \frac{y_i + h_i}{H_I}, \frac{A_i}{A_I})$, where ($x_i$, $y_i$) are the center coordinates, ($w_i$, $h_i$) are the width and height, $A_i$ and $A_I$ are the areas of region $i$ and image $I$, and $W_I$ and $H_I$ are the width and height of the image $I$.
To represent union boxes, we compute the following 9-d feature:
\begin{align}
    l_{s\cup o} = \bigg( 
    &\frac{x_s - x_o}{w_o}, 
    \frac{y_s - y_o}{h_o}, \log\frac{w_s}{w_o}, \log\frac{h_s}{h_o}, \nonumber \\  & 
    \frac{x_o - x_s}{w_s}, 
    \frac{y_o - y_s}{h_s}, \log\frac{w_o}{w_s}, \log\frac{h_o}{h_s}, \frac{A_u}{A_I} \bigg) , \nonumber
\end{align} 
where ($x_s$, $y_s$, $w_s$, $h_s$) and ($x_o$, $y_o$, $w_o$, $h_o$) are the subject and object box coordinates and $A_u$ is the area of the union box. 
In our network, all location features $(l_s, l_o, l_{s\cup o})$ are first concatenated into a 19-d vector, which is then fed into a two-layer MLP with intermediate layer dimenension of 32 and output dimension of 16.

\smallskip
\noindent{\bf UVTransE Module.} 
In this stage, each pair of objects, together with their union features, 
are sent to UVTransE, which is discussed in detail in Section~\ref{sec:uvtranse}. 
After performing UVTransE, the outputs are passed through two FC layers of input-output sizes of 256 (appearance) + 16 (location) $\rightarrow$ 256, and 256 $\rightarrow |P|$  to produce a confidence score per predicate. 
These scores can be used as-is to output a set of ranked relationships, or can be combined with the scores of the language module.

\smallskip
\noindent{\bf Language Module.}
As stated in Section \ref{sec:uvtranse}, our language module is based on  bi-GRUs~\cite{Schuster:1997:BRN:2198065.2205129}. We use GloVe~\cite{pennington2014glove} for our word embedding to encode subject and object class names. Then we get the predicate embedding $\hat{\boldsymbol{p}}$ from UVTransE (Eq. \ref{eq:hatp}) and put it through a fully connected (FC) layer to get the same dimensionality as GloVe. Next, we feed the subject, predicate, and object encodings into three successive steps of a bi-directional GRU (Bi-GRU)~\cite{Schuster:1997:BRN:2198065.2205129}. 
The hidden states, which are 100-d, are then concatenated across the three time steps and both directions are used for predicate classification with two FC layers of size $600 \rightarrow 256 \rightarrow |P|$.



\section{Experiments}
\label{sec:exp}

In Section \ref{subsec:vrd}, we begin by evaluating our method on the VRD dataset~\cite{lu2016visual}, which is moderate in size and is one of the most common benchmarks for relationship detection. Because we are especially interested in the setting of rare and unusual relations, Section \ref{subsec:unrel} presents an evaluation on the UnRel dataset~\cite{Peyre17}, which is small and can only be used for testing. Finally, to demonstrate that our method also works well on larger-scale benchmarks, as well as on the recently introduced task of scene graph generation, Sections \ref{subsec:vg} and \ref{subsec:openimages} report results on two subsets of the Visual Genome~\cite{Krishna:2017:VGC:3088990.3089101} and Google's Open Images~\cite{OpenImages}. 

\subsection{{\bf Results on the Stanford VRD Dataset}}
\label{subsec:vrd}
\noindent{\bf Dataset.} 
We follow the methodology of~\cite{lu2016visual} to evaluate our method on the Stanford VRD dataset~\cite{lu2016visual}. This dataset contains 5,000 images with 100 object categories and 70 predicates. It has around 30k relation annotations, with an average of 8 relations per image. We use the same train/test split as in~\cite{lu2016visual}, consisting of 4,000 training images and 1,000 test images. In this specific split, 1,877 relationships in the test set never occur in the training set, thus allowing us to evaluate zero-shot prediction. 


\smallskip
\noindent{\bf Dataset-specific details.}
We use Faster R-CNN with the VGG-16 backbone to obtain candidate objects. The VGG-16 network is initialized with parameters pre-trained on ImageNet and fine-tuned on the VRD dataset and a subset of Visual Genome. Specifically, because some objects have less than 50 instances in the VRD training set, we take at least 500 instances for each class from Visual Genome. Our object detector has an mAP of 19.1 on the VRD dataset. This is low in absolute terms, partly due to incomplete ground truth annotations, but higher than the 13.98 mAP reported by Zhang et al.~\cite{Zhang_2017_CVPR}. To obtain subject and object boxes for training UVTransE, we use ground truth boxes as well as detected boxes with $IoU \ge 0.5$. At test time, for each image, we use the top 30 candidate object boxes returned by Faster R-CNN for mining relationships. 

We freeze the weights of the detector while jointly training UVTransE and language modules. The hyper-parameters used for the VRD dataset are $C = 1.0$ (regularization constant, Eq.~\ref{eq:C}) and $\alpha = 0.5$ (visual-language weighting, Eq.~\ref{eq:alpha}). SGD is used as the optimizer with an initial learning rate of $1e^{-3}$ for the detector, UVTransE, and the language module.

\smallskip
\noindent{\bf Evaluation metrics.}
Our evaluation methodology is consistent with~\cite{lu2016visual}.
Given a test image, the VRD model being evaluated is used to score all possible predicates between every pair of detected objects, retaining only the top $k$ best-scoring predicates for each pair. Then we rank all these predictions and report {\bf Recall@50} and {\bf Recall@100}, or the fraction of ground-truth triplets that are correctly recalled in the top 50 or 100. The evaluation is done for three setups. 
\begin{enumerate}
    \item {\bf Predicate detection:} To investigate whether the VRD model is good at detecting relations, independent of the quality of object detection, we measure the accuracy of predicate prediction when the ground truth object classes and boxes are given. A few previous works~\cite{dai2017detecting, liang2018Visual} evaluate their predicate detection under the $k = 70$ setting, where $k$ is the number of chosen predicates for each object pair, to achieve better recall. However, we stick to the original setting~\cite{lu2016visual} and evaluate it for $k = 1$.
    \item {\bf Phrase detection:} In this setting, a prediction is considered correct if a triplet $(s, r, o)$ is correctly recognized, and the area of intersection over union (IoU) between the predicted $s \cup o$ box and the ground-truth is above 0.5.
    \item {\bf Relationship detection:} This is similar to phrase detection, except that it requires the IoU for subject and object box to both be above 0.5, which is more challenging.
\end{enumerate}

\begin{table}[t!]
 \begin{center}
  \resizebox{\columnwidth}{!}{
  \begin{tabular}{ l c c c c}
   \hline
   \multicolumn{1}{l}{} & \multicolumn{2}{c}{All Test} & \multicolumn{2}{c}{Zero-shot Only}\\
   & Phr. Det. & Rel. Det.& Phr. Det. & Rel. Det. \\
   \hline
   {\bf $C = 0$} & 6.48 & 4.67 & 4.28 & 3.08 \\
   {\bf $C = 0.5$} & 22.14 & 18.96 & 11.21 & 9.75\\
   {\bf $C = 1.0$} & {\bf 23.92} & {\bf 20.22} & {\bf 11.77} & {\bf 10.21}\\
   {\bf $C = 1.5$} & 23.38 & 19.99 & 11.46 & 9.75 \\\hline
  \end{tabular}
  }
 \end{center}
 \caption{
 The effect of $C$ on Recall@50 on the Stanford VRD dataset. {\bf Bold} indicates highest numbers.} 
 \label{table:vrd_ablation_C}
 \vspace{-5pt}
\end{table}

\begin{table*}[b!]
 \begin{center}
  \begin{tabular}{ p{3.5cm} c c | c c c c | c c c c}
   \hline
   \multicolumn{1}{p{3.5cm}}{} &
   \multicolumn{2}{c}{Predicate Det.} &
   \multicolumn{4}{c}{Phrase Det.} & \multicolumn{4}{c}{Relationship Det.}\\
   \multicolumn{1}{p{3cm}}{} &
   \multicolumn{1}{c}{All} &
   \multicolumn{1}{c}{Zero-shot} &
   \multicolumn{2}{c}{All} &
   \multicolumn{2}{c}{Zero-shot} &
   \multicolumn{2}{c}{All} &
   \multicolumn{2}{c}{Zero-shot}\\
   &R@50 &R@50& R@50 &R@100& R@50 &R@100& R@50 &R@100& R@50 &R@100 \\
   \hline
   {\bf Appearance} & 18.17 & 7.44 & 8.59 & 10.68 & 5.34 & 10.11 & 7.52 & 9.11 & 4.82 & 8.97 \\
   {\bf Appearance + spatial} & 38.89 & 14.35 & 20.06 & 24.70 & 7.98 & 11.84 & 17.02 & 20.54 & 6.90 & 10.02\\
   {\bf Summation} & 49.01 & 18.52 & 21.93 & 27.80 & 10.25 & 14.94 & 17.78 & 21.37 & 9.47 & 13.33\\
   {\bf VTransE [V] (our impl.)} & 45.12 & 12.84 & 19.74 & 25.62 & 7.27 & 10.61 & 16.21 & 20.48 & 6.31 & 9.55 \\
  {\bf VTransE [V+L] (our impl.)} & 50.11 & 15.31 & 26.13 & 31.40 & 8.73 & 12.05 & 22.23 & 26.14 & 7.67 & 10.99 \\
   {\bf UVTransE [V]} & 49.98 & 22.92 & 23.92 & 29.57 & 11.77 & 17.41 & 20.22 & 24.13 & 10.21 & 15.92\\
   {\bf UVTransE [V+L]} & {\bf 55.46} & {\bf 26.49} & {\bf 30.01} & {\bf 36.18} & {\bf 13.07} & {\bf 18.44} & {\bf 25.66} & {\bf 29.71} & {\bf 11.00} & {\bf 16.78} \\\hline
  \end{tabular}
 \end{center}
\caption{Comparisons of baselines to our proposed method on the Stanford VRD dataset. {\bf Bold} indicates highest numbers.}
\label{table:vrd_ablation}
\end{table*}


\smallskip
\noindent{\bf Ablation Study.}
First, we perform ablation studies to evaluate the effectiveness of different components of our model. Table~\ref{table:vrd_ablation_C} shows the performance of our model for different values of the regularization parameter $C$ on the embedding weights (Eq.~\ref{eq:C}). The low performance for $C=0$ confirms that regularizing the norms of projected subject, object, and predicate vectors is important for learning effective embeddings in our framework. The value of $C=1$ gives us the best results so we use it in all subsequent experiments.

In Table~\ref{table:vrd_ablation}, we compare our method to several baselines using the same trained detector, and thus, the same predicted bounding boxes, detector confidence scores, and visual features to describe the boxes. The simplest baseline, called {\bf Appearance}, is to directly classify the predicate based on the concatenated visual features of the subject, object, and union boxes. The second baseline, {\bf Appearance + spatial}, concatenates spatial features described in Section \ref{sec:overall_pipeline} with the appearance features. Both methods learn a single projection matrix, but use the same weight regularization as described in Section \ref{sec:uvtranse}. The results confirm that adding spatial features to purely appearance-based features significantly improves performance. 
The third baseline, {\bf Summation}, uses summation instead of subtraction in Eq. (\ref{eq:hatp}). Since this formulation is very similar to UVTransE, we run it to validate the effectiveness of the subtractive model. Across all our metrics, the results for Summation are 1-2\% below those of UVTransE. This shows that despite the superficial similarity, the subtractive model better captures the structure of the VRD problem.  
The baseline in the fourth line of Table~\ref{table:vrd_ablation} is our own re-implementation of {\bf VTransE}~\cite{Zhang_2017_CVPR}, for which we found that we had to add our regularization terms (Eq. \ref{eq:C}) to achieve results comparable to~\cite{Zhang_2017_CVPR}. We show the performance of two variants: without a language model ({\bf UVTransE [V]}), or with our Bi-GRU language model ({\bf UVTransE [V+L]}). Both variants include our spatial features. Compared to {\bf VTransE [V]}, {\bf UVTransE [V]} boosts performance significantly both in the general and in the zero-shot case. For the predicate detection task, the absolute improvement is about 5\% in the general case and 10\% in the zero-shot case, confirming that incorporating the union box in the translational formulation helps to isolate predicate information, particularly for rare and previously unseen cases.
Adding the language module benefits {\bf VTransE} and {\bf UVTransE} about the same, although the absolute improvements are smaller in the zero-shot case than in the general case ($\sim2$ - $3\%$ vs. $\sim5\%$)  as the language model tends to bias predictions towards relationships seen during training. 

\begin{table*}[t!]
 \begin{center}
  \resizebox{\textwidth}{!}{
  \begin{tabular}{ p{7.5cm} l c c c c c c}
   \hline
   \multicolumn{1}{p{7.5cm}}{} & Detector (pre-training) & mAP & ROI feature & Spatial feature & Language feature & Joint reasoning & Extra training data \\\hline
   Visual Relationship Detection ({\bf VLK}, 2016) \cite{lu2016visual} & VGG &  &  & - & \checkmark & - & - \\
   Visual Translation Embedding ({\bf VTransE}, 2017) \cite{Zhang_2017_CVPR} & VGG & 13.98 & & \checkmark & - & - & -\\
   Variation-Structured Reinforcement Learning ({\bf VRL}, 2017) \cite{vrl17} & VGG (ImageNet) &  &  &- & \checkmark & \checkmark & -\\
   Weakly-supervised learning of visual relations ({\bf SA-full}, 2017) \cite{Peyre17} & VGG (ImageNet) &  &  & \checkmark & - & - & -\\
   Canonical Correlation Analysis ({\bf CCA}, 2017) \cite{plummerPLCLC2017} & VGG (COCO) &  &  & \checkmark & \checkmark & - & -\\
   Linguistic Knowledge Distillation ({\bf LK}, 2017) \cite{yu2017_vrd_knowledge_distillation} & VGG &  &  & \checkmark & \checkmark & - & Wikipedia \\
   Context-aware Interaction Recognition ({\bf CAIR}, 2017)~\cite{zhuang2017towards} & VGG &  &  & \checkmark & \checkmark & - & - \\
   {\bf Zoom-Net} (2018) \cite{gjyin_eccv2018} & VGG &  &  & \checkmark & - & \checkmark & - \\
   Relationship Detection Network  ({\bf RelDN}, 2019) \cite{zhang2019vrd} & VGG (COCO) &  & Align & \checkmark & \checkmark & - & -\\
   Large Scale Visual Relationship ({\bf LS-VRD}, 2019)~\cite{zhang2018large} & VGG (COCO) &  &  & - & \checkmark & - & - \\
   \hline
   Deep Relational Networks ({\bf DR-Net}, 2017) \cite{dai2017detecting} & VGG (ImageNet) &  &  & \checkmark & - & - & - \\
   Deep Structural Ranking ({\bf DSR}, 2018) \cite{liang2018Visual} & VGG &  &  & \checkmark & \checkmark & - & - \\\hline
   {\bf UVTransE [V+L] } & VGG (ImageNet) & 19.10 & Align & \checkmark & \checkmark & - & Visual Genome \\\hline
  \end{tabular}
  }
 \end{center}
 \caption{
Summary of state-of-the-art methods on the VRD dataset. The `Detector' column lists the architecture of the detector and the dataset used for pre-training (if mentioned in the original paper). `ROI feature' indicates the type of ROI feature used (in papers that do not explicitly mention using ROIAlign, we assume ROIPool is used). `mAP' lists the accuracy of the detector. `Spatial feature' and `language feature' indicate whether bounding box features similar to the ones of Section \ref{sec:overall_pipeline} and a language model similar to the one of Section \ref{subsec:language_module} are used. `Joint reasoning' indicates whether the method uses context or joint reasoning instead of predicting each pairwise relationship separately. `Extra training data' indicates whether additional data is used for training either the detector or the language model. In each column, \checkmark indicates the presence of features, - indicates absence, and blank means the information is not provided in the original paper.} 
 \label{table:vrd_model_comparison}
\end{table*}

\begin{table*}[tp!]
 \begin{center}
 \resizebox{\textwidth}{!}{
  \begin{tabular}{ l c | c c c c c c | c c c c c c }
   \hline
   \multicolumn{1}{l}{} &
   \multicolumn{1}{l}{Predicate Det.} &
   \multicolumn{6}{c}{Phrase Det.} & \multicolumn{6}{c}{Relationship Det.}\\
   \multicolumn{1}{l}{} &
   \multicolumn{1}{c}{} &
   \multicolumn{2}{c}{$k = 1$} &
   \multicolumn{2}{c}{$k = 10$} &
   \multicolumn{2}{c}{$k = 70$} &
   \multicolumn{2}{c}{$k = 1$} &
   \multicolumn{2}{c}{$k = 10$} &
   \multicolumn{2}{c}{$k = 70$}\\
   &R@50 &R@50& R@100 &R@50& R@100 &R@50& R@100 &R@50& R@100 &R@50& R@100 &R@50& R@100 \\
   \hline
   {\bf VLK} \cite{lu2016visual} & 47.87 & 16.17 & 17.03 & - & - & - & - & 13.86 & 14.07 & - & - & - & -\\
   {\bf VTransE} \cite{Zhang_2017_CVPR} & 44.76 & 19.42 & 22.42 & - & - & - & - & 14.07 & 15.20 & - & - & - & - \\
   {\bf VRL} \cite{vrl17} & - & 21.37 & 22.60 &  - & - & - & - & 18.19 & 20.79 & - & - & - & -\\
   {\bf SA-full} \cite{Peyre17} & 50.40 & 16.70 & 18.10 & - & - & - & - & 14.90 & 16.10 & - & - & - & - \\
   {\bf CCA} \cite{plummerPLCLC2017} & - & - &  - & 16.89 & 20.70 & - & - & - & - & 15.08 & 18.37 & - & - \\
   {\bf LK} \cite{yu2017_vrd_knowledge_distillation}$^{\color{red}\bigoasterisk}$ & \underline{55.16} & 23.14 &  24.03 & 26.47 & 29.76 & 26.32 & 29.43 & 19.17 & 21.34 & 22.56 & 29.89 & 22.68 & 31.89 \\
   {\bf CAIR} \cite{zhuang2017towards} & - & 24.04 & 25.56 & - & - & - & - & 20.35 & 23.52 & - & - & - & -\\
   {\bf Zoom-Net} \cite{gjyin_eccv2018} & 50.69 & 24.82 & 28.09 & - & - & 29.05 & 37.34 & 18.92 & 21.41 & - & - & 21.37 & 27.30 \\
   {\bf RelDN} \cite{zhang2019vrd} & - & {\bf 31.34} & {\bf 36.42} & {\bf 34.45} & {\bf 42.12} & {\bf 34.45} & {\bf 42.12} &  \underline{25.29} & \underline{28.62} & {\bf 28.15} & \underline{33.91} & {\bf 28.15} & \underline{33.91}\\
   {\bf LS-VRD} \cite{zhang2018large} & - & 28.93 & 32.85 & \underline{32.90} & 39.66 & \underline{32.90} & 39.64 & 23.68 & 26.67 & 26.98 & 32.63 & 26.98 & 32.59 \\
   \hline
   {\bf DR-Net} \cite{dai2017detecting}$^{\color{blue}\bigstar}$ & 80.78 & - & - & - & - & 19.93 & 23.45 & - & - & - & - & 17.73 & 20.88  \\
   {\bf DSR} \cite{liang2018Visual}$^{\color{blue}\bigstar}$ & 86.01 & - & - & - & - & - & - & - & - & - & - & 19.03 & 23.29 \\\hline
   {\bf UVTransE [V+L] } & {\bf 55.46} & \underline{30.01} & \underline{36.18} & 31.82 &  \underline{40.43} &  31.51 & \underline{39.79} & {\bf25.66} & {\bf 29.71} &  \underline{27.41} & {\bf 34.55} & \underline{27.32} & {\bf 34.11}   \\\hline
  \end{tabular}
  }
 \end{center}
\caption{Full test set performance on the Stanford VRD dataset. {\bf Bold} indicates highest numbers, \underline{underline} indicates second-highest.
$\color{red}\bigoasterisk$ indicates use of large-scale external Wikipedia data. 
$\color{blue}\bigstar$ indicates $k=70$, instead of $k=1$ for predicate detection (See Sec.~\ref{subsec:vrd}).}
\label{table:vrd_seen}
\vspace{4pt}
\end{table*}

\begin{table*}[tp!]
 \begin{center}
  \begin{tabular}{ l c c c c c}
   \hline
   \multicolumn{1}{l}{} & Predicate Det. & \multicolumn{2}{c}{Phrase Det.} & \multicolumn{2}{c}{Relationship Det.}\\
   & R@50 &R@50& R@100 &R@50& R@100 \\
   \hline
   {\bf VLK} \cite{lu2016visual} & 8.45 & 3.36 & 3.75 & 3.13& 3.52\\
   {\bf VTransE} \cite{Zhang_2017_CVPR} \cite{Zhang_2017_CVPR} & - & 2.65 & 3.51 & 1.71 & 2.14\\
   {\bf VRL} \cite{vrl17} & - & 9.17 & 10.31 & 7.94 & 8.52\\
   {\bf SA-full} \cite{Peyre17} & \underline{23.60} & 7.4 & 8.7 & 7.1 & 8.2\\
   {\bf CCA} \cite{plummerPLCLC2017} & - & {10.86} & {15.23}  & {9.67} & {13.43} \\
   {\bf LK} \cite{yu2017_vrd_knowledge_distillation}$^{\color{red}\bigoasterisk}$ & {16.98} & \underline{13.01} & \underline{17.24} & \textbf{12.31} & \underline{16.15}\\
   {\bf CAIR} \cite{zhuang2017towards} & - & 10.78 & 11.30 & 9.54 & 10.26 \\
   {\bf Zoom-Net } \cite{gjyin_eccv2018} & - & - & - & - & - \\
   {\bf RelDN} \cite{zhang2019vrd} & - & - & - & - & - \\
   {\bf LS-VRD} \cite{zhang2018large} & - & - & - & - & - \\
   \hline
   {\bf DR-Net} \cite{dai2017detecting} & - & - & - & - & - \\
   {\bf DSR} \cite{liang2018Visual}$^{\color{blue}\bigstar}$ & 60.90 & - & - & 5.25 & 9.20 \\\hline
   {\bf UVTransE [V+L] } &  {\bf 26.49} & {\bf 13.07} & {\bf 18.44} & \underline{11.00} & {\bf 16.78} \\\hline
  \end{tabular}
 \end{center}
 \caption{Zero-shot performance on the Stanford VRD dataset. {\bf Bold} indicates highest numbers, \underline{underline} indicates second-highest.
 $\color{red}\bigoasterisk$ indicates use of large-scale external Wikipedia data. $\color{blue}\bigstar$ indicates $k=70$, instead of $k=1$ for predicate detection (See Sec.~\ref{subsec:vrd}).
 We treat $k$ as a hyper-parameter that can be cross validated for phrase and relationship detection. In our case, $k = 10$.} 
 \label{table:vrd_unseen}
\end{table*}

\smallskip
\noindent{\bf Comparison to the state of the art.}
Next, we compare performance to an extensive collection of VRD models from the recent literature, which are summarized in Table \ref{table:vrd_model_comparison}.  {\bf VLK}~\cite{lu2016visual} is a two-stage model that uses both appearance features and language priors for relationship prediction. {\bf VTransE}~\cite{Zhang_2017_CVPR} is the main method we build upon, as discussed previously. Note that the results reported by~\cite{Zhang_2017_CVPR} differ from those of our re-implementation discussed above due to the use of different detectors and our inclusion of norm regularization in the training objective. {\bf VRL}~\cite{vrl17} applies a deep variation-structured reinforcement learning framework to sequentially discover object relationships and attributes using appearance and language features. {\bf SA-full}, the fully supervised version of the method of Peyre et al.~\cite{Peyre17}, uses appearance and spatial features to handle multi-modal relations and generalize well to unseen triplets.
{\bf DR-Net}~\cite{dai2017detecting} exploits statistical dependencies between objects and their relationships when modeling relations.
{\bf DSR}~\cite{liang2018Visual} designs a ranking objective that enforces the annotated relationships to have higher scores than negative examples.
{\bf CCA}~\cite{plummerPLCLC2017} utilizes multiple CCA embedding cues (both vision and language), along with an SVM for ranking relationship proposals. {\bf LK}~\cite{yu2017_vrd_knowledge_distillation} distills large-scale external linguistic knowledge from Wikipedia to achieve better performance for rare relationships. {\bf CAIR}~\cite{zhuang2017towards} builds one classifier for each predicate, but the classifier parameters are also adaptive to the context, i.e.\  (\emph{subject}, \emph{object}) pairs. {\bf Zoom-Net} ~\cite{gjyin_eccv2018} encourages deep message interactions between local object features and global predicate features to recognize relationships. 
{\bf RelDN}~\cite{zhang2019vrd} is one of the most recent methods that achieves state-of-the-art performance using graphical contrastive losses to better learn subtle subject-object associations.
Finally, {\bf LS-VRD}~\cite{zhang2018large}  learns a visual and a semantic module that map features from the two modalities into a shared space such that the relations are discriminative.

It must be stated that getting completely apples-to-apples comparisons against the above methods is difficult as they vary in a number of respects. Among the most important is the quality of the underlying object detector, which depends on the network architecture and training protocol (details of training are usually not fully discussed in the papers, nor are the accuracies of the detector always reported). Other factors include the feature descriptors (in particular, whether spatial or linguistic features are included), the use of external data for training detectors or language model, the type of inference performed, the evaluation protocol, and so on. In an attempt to be transparent about these sources of variation, we list them in Table \ref{table:vrd_model_comparison}. With these caveats in mind, Table~\ref{table:vrd_seen} compares our results to published numbers from the above papers the full VRD test set. Our model with the visual feature alone, UVTransE [V], reaches comparable performance to CAIR and Zoom-Net. After including the language module, we outperform all methods except for the most recent RelDN (which uses a different detector pre-trained on COCO~\cite{lin2014microsoft}). 

Table~\ref{table:vrd_unseen} presents a comparative evaluation for the zero-shot setting. Our method surpasses all other methods that use only the given dataset for training, and is comparable to LK, which incorporates external language data. Significantly, several of the strongest methods from Table \ref{table:vrd_seen}, including  Zoom-Net, RelDN, and DR-Net, do not report their results for the zero-shot setting at all. At least in some cases, this is because achieving high performance on common relations comes at the cost of very low performance on rare relations. In particular, we tested the RelDN model published by the authors~\cite{zhang2019vrd} on the zero-shot test set and obtained accuracies close to 0 on all metrics, with almost all rare relationships being confidently classified as `no relationship'. 

\smallskip

\begin{figure*}[t]
\centering
    \includegraphics[trim={1cm 0.8cm 1cm 0.8cm}, width=0.8\textwidth,height=0.3\textheight]{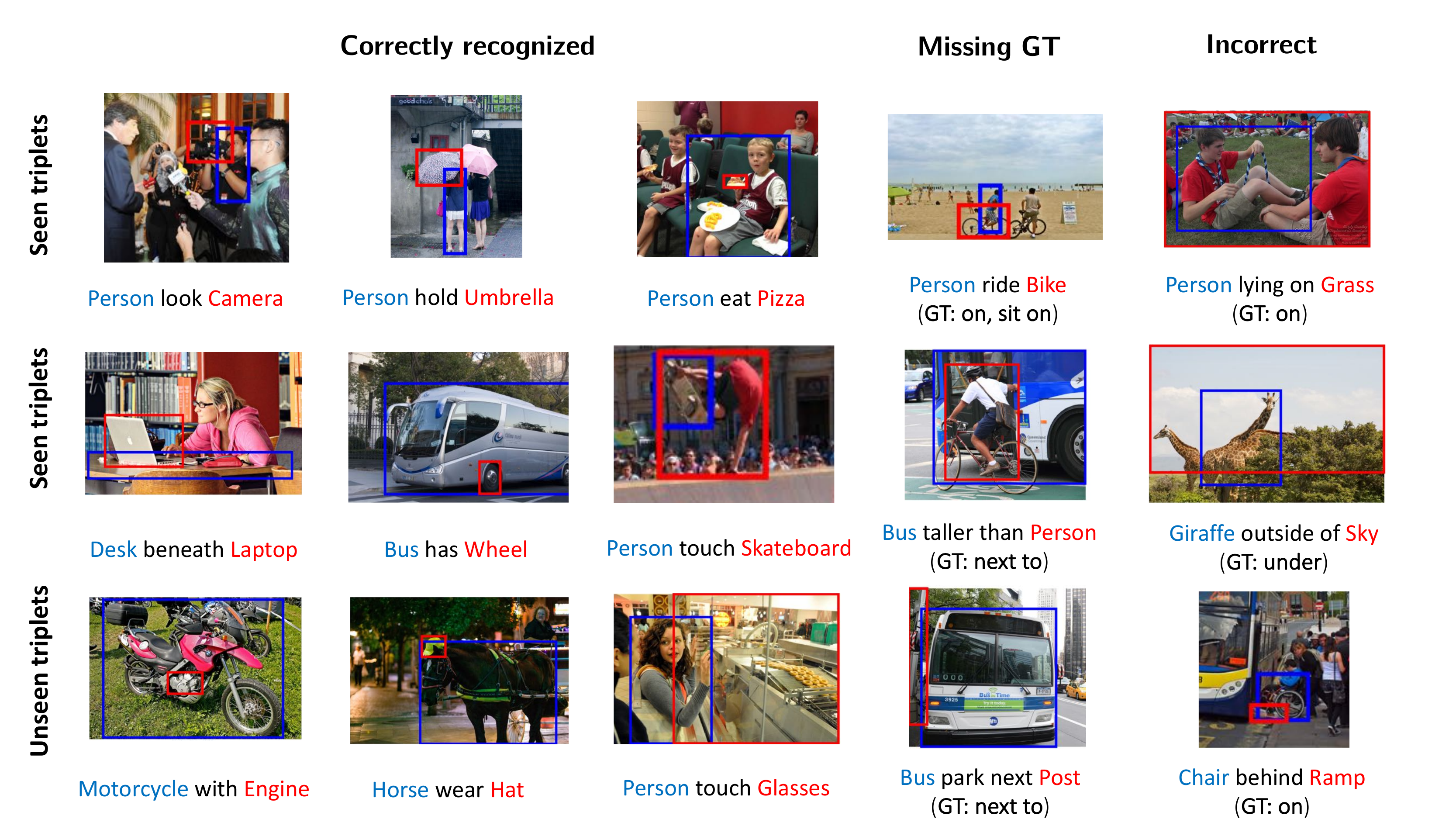}
   \caption{Examples of relationship detection on the VRD test split. A triplet is correctly recognized if both the bounding boxes are correctly localized and the predicate matches the ground truth. The `Missing GT' column shows relationships that were marked as incorrect since they are not present in the ground truth. The `Incorrect' column shows legitimate mistakes. The last row shows our zero-shot results.}
\label{fig:vrd_res}
\end{figure*}

\noindent {\bf Qualitative results.} Figure~\ref{fig:vrd_res} shows example predictions by our model for both seen and unseen relationships. There are many plausible detected triplets that are marked as negatives due to the lack of annotations (Missing GT column). In some cases, predicates are not mutually exclusive. For example, (\emph{person}, \emph{on}, \emph{bike}) can also be labeled as (\emph{person}, \emph{ride}, \emph{bike}); however, predicting \emph{ride} for this pair of objects is penalized due to the missing ground truth.


\subsection{{\bf Results on the UnRel Dataset}} \label{subsec:unrel}
\noindent{\bf Dataset.} 
To further investigate the generalization ability of our method, we perform experiments on the UnRel dataset~\cite{Peyre17}. It consists of 1071 images with 76 unusual triplets, such as (\emph{person}, \emph{ride}, \emph{dog}), (\emph{car}, \emph{under}, \emph{elephant}), etc. The ground truth on this dataset is more exhaustively annotated than on VRD. 

UnRel contains too few images for training, so we simply use it as a test set for ourUVTransE model trained on the VRD dataset. The hyperparameters are the same as in Section \ref{subsec:vrd}.

\smallskip
\noindent{\bf Evaluation metrics.}
Following \cite{Peyre17}, we evaluate retrieval and localization with mAP over triplet queries $(\emph{s}, \emph{p}, \emph{o})$ in two settings: 
\begin{enumerate}
\item {\bf With ground truth:} We are given GT pairs of boxes ($b_s$, $b_o$) and then rank them based on their predicate scores $z_p$ (Eq.~\ref{eq:z_p}). The purpose of this setup is to test the ``predicate prediction'' part only, without the contribution of the object detector. 
\item {\bf With candidates:} Candidate boxes ($b_s$, $b_o$) are provided by the object detector and ranked according to the combined score $z_{(s, p, o)}$ (Eq.~\ref{eq:vis_only_score}). In this setting, we also have to evaluate the accuracy of localization. According to \cite{Peyre17}, a candidate pair of boxes is positive if its IoU with GT pair is above 0.3. There are three localization metrics: {\bf mAP-subj}: the subject box itself should have at least 0.3 overlap with its GT; {\bf mAP-union}: the entire relationship is localized as one bounding box and it should have at least 0.3 overlap with the GT; {\bf mAP-subj/obj}: Both subject and object boxes should have at least 0.3 overlap with their corresponding GT boxes.
\end{enumerate}

\smallskip
\noindent{\bf Comparison with state of the art.}
We compare our results with numbers from four methods reported by~\cite{Peyre17}.
The {\bf chance} baseline randomly orders the proposals.
The second method is {\bf DenseCap}~\cite{densecap}, where the output bounding box is interpreted as either a subject box or a union box for evaluation, as suggested in~\cite{Peyre17}.
{\bf VLK}~ \cite{lu2016visual} is the result from the re-implementation of~\cite{lu2016visual} by \cite{Peyre17}. Finally, {\bf SA-full}~\cite{Peyre17} is, to our knowledge, the state-of-the-art fully supervised method on UnRel. 
As previously mentioned, our model is only trained on the Stanford VRD dataset, and is evaluated on the UnRel dataset without any changes, similar to the VLK and SA-full methods.
\begin{table}[tp!]
\begin{center}
\resizebox{\columnwidth}{!}{%
  \begin{tabular}{ p{2.7cm} c c c c}
   \hline
    & \multirow{2}{*}{With GT} & \multicolumn{3}{c}{With candidates} \\
    & & union & subj & subj/obj \\
   \hline
   {\bf Chance} & 38.4 & 8.6 & 6.6 & 4.2 \\
   {\bf DenseCap} \cite{densecap} & - & 6.2 & 6.8 & - \\
   {\bf VLK} \cite{lu2016visual} & 50.6 & 12.0 & 10.0 & 7.2 \\
   {\bf SA-full} \cite{Peyre17} & 62.6 & 14.1  & 12.1 & 9.9\\\hline
   {\bf UVTransE [V]} & 70.6 & {\bf 19.2} & {\bf 17.2} & \textbf{14.8} \\
   {\bf UVTransE [V+L]} & \textbf{71.7} & 18.0 &  16.3 & 14.1 \\
   \hline
  \end{tabular}
  }
 \end{center}
 \caption{Retrieval on UnRel (mAP) with IoU=0.3.} \label{table:unrel}
\end{table}

The retrieval results in Table~\ref{table:unrel} show that our model consistently outperforms all other methods. 
Interestingly, our language module improves the accuracy when the ground truth boxes are given, but degrades it slightly when the objects are provided by the object detector, likely because the language model gets confused if the predicted object classes are wrong, or the boxes are incorrectly localized. This behavior is different from what we observed on zero-shot evaluation for the VRD dataset, since the images in UnRel are deliberately unusual and hard. 

\begin{figure*}[t]
\centering
    \includegraphics[
    trim={1cm 1.2cm 0 2cm},clip,width=0.8\textwidth,height=0.25\textheight]{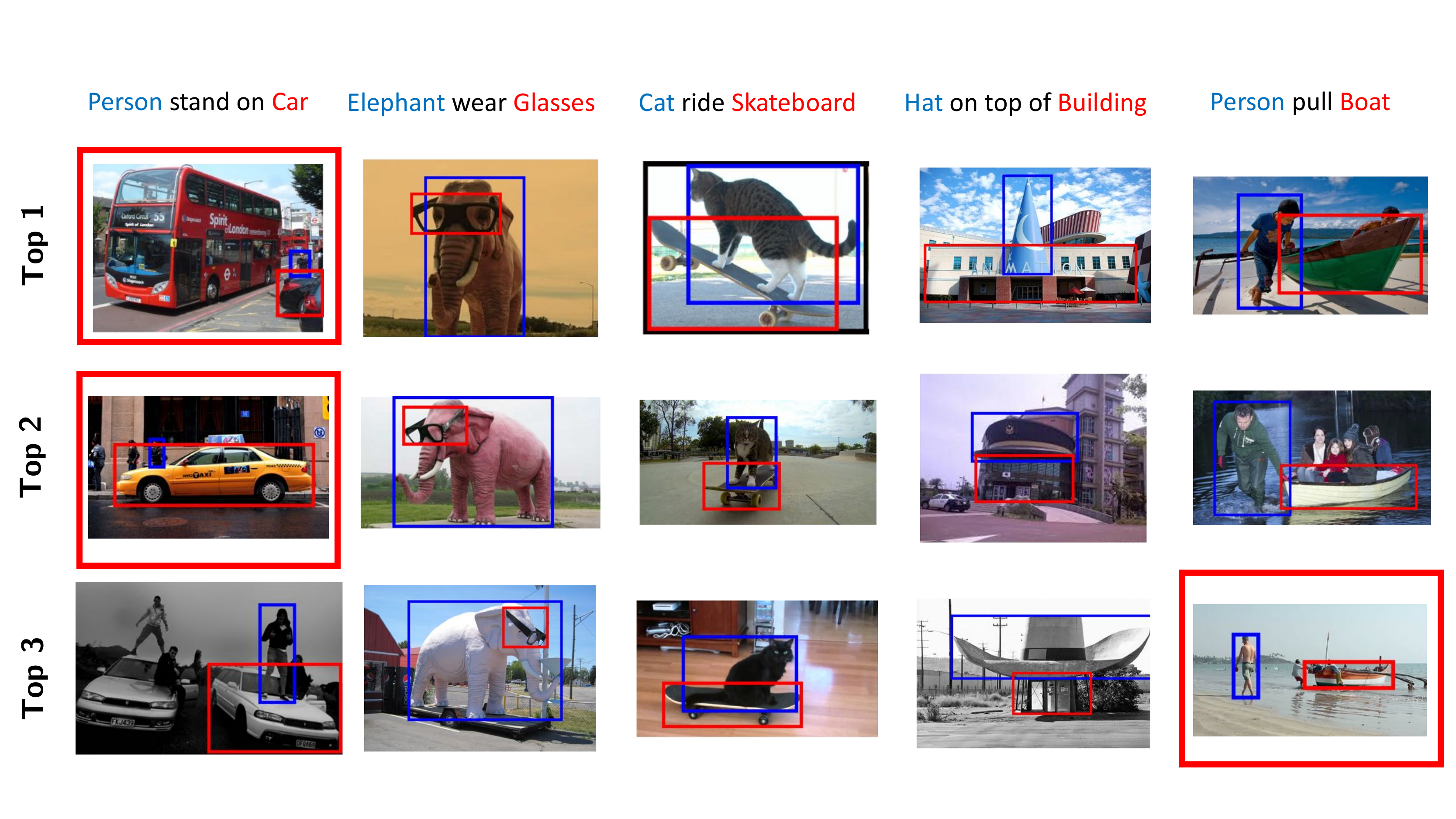}
   \caption{Top three retrievals for a set of UnRel triplet queries with our model. A relationship is marked as positive if the subject and object boxes have $IoU \geq 0.3$ with the ground truth. Otherwise, it is marked as an error (red box around the entire image).}
\label{fig:unrel_res}
\end{figure*}

Figure~\ref{fig:unrel_res} shows the top triplets retrieved by our model for some representative queries.
We use red boxes around images to indicate wrongly retrieved examples. It can be seen that we are able to successfully retrieve examples of rare relations such as (\emph{elephant}, \emph{wear}, \emph{glasses}), and  (\emph{hat}, \emph{on top of}, \emph{building}).

\subsection{{\bf Results on the Visual Genome Dataset}} \label{subsec:vg}
\noindent{\bf Datasets.} 
To demonstrate the effectiveness of our model on large-scale datasets that are more geared towards scene graph generation, we perform experiments on two cleaned subsets of Visual Genome~\cite{Krishna:2017:VGC:3088990.3089101}.  The first one, created by Xu et al.~\cite{xu2017scenegraph}, is composed of the most frequent 
150 objects and 50 predicates. We call this one {\bf VG-IMP} after the method of~\cite{xu2017scenegraph}. After pre-processing, VG-IMP is split into training and test sets containing 75,651 images and 32,422 images, respectively.
The second subset, created by Zhang et al.~\cite{Zhang_2017_CVPR}, contains an even larger number of objects and predicates, 200 and 100, respectively. We follow the same 73,801/25,857 train/test split as in \cite{Zhang_2017_CVPR}. We call this subset {\bf VG-VTransE}.

\smallskip
\noindent{\bf Implementation details.}
\label{sec:vg_setup}
On the VG-IMP subset, we train a Faster R-CNN detector with a VGG-16 backbone to obtain an mAP of 19.2\%. Competing methods using the same backbone report higher performance, namely,  20.0\% for Neural Motifs~\cite{zellers2018scenegraphs}, 20.4\% for Graph R-CNN~\cite{jwyang_graph_rcnn}, and 25.5\% for RelDN~\cite{zhang2019vrd}. Maximizing the accuracy of object detection is not the focus of our work, but to better compete with these methods on the final accuracies for relationship localization, we trained a stronger detector using a ResNet-101 backbone, for an mAP of 23.8\%. For the VG-VTransE subset, the only competing method with published results is VTransE~\cite{Zhang_2017_CVPR}, which uses a VGG-16 backbone. On that subset, the mAP of our VGG-16 detector is 12.5\%, which is sufficient to compete with~\cite{Zhang_2017_CVPR}.  We initialize the parameters in Faster R-CNN with ImageNet pre-training. After fine-tuning the parameters on the respective Visual Genome subset, we fix it, and train our UVTransE module along with the language module with initial learning rate of $1e^{-2}$. At test time, for each image, we use the top 50 candidate object proposals ranked by Faster R-CNN for mining relationships. 

To get good performance on Visual Genome evaluation metrics (described below), we found it useful to add a `background' or `no relationship' class during training. We define positive relation triplets as those where both subject and object have $IoU \geq 0.5$. During training, for each image, we sample 32 relations with the ratio of positive to negative triplets being $1:3$. On the VRD dataset (Section \ref{subsec:vrd}), this kind of sampling improves performance for common relationships, but significantly degrades performance for the zero-shot case, as many unseen relationships get classified as `background' with high confidence. 

For the results of this section, we also found it necessary to change Eqs. (\ref{eq:vis_only_score}) and (\ref{eq:alpha}) to use product instead of addition: 
\begin{equation}
z_{\left(s, p, o\right)} = z_s \times z_o \times z_{p} \,,
\label{eq:vis_only_score_product}
\end{equation}
and for {\bf UVTransE [V+L]},
\begin{align}
z_{\left(s, p, o\right)} &= (\alpha z_{p} + (1 - \alpha) z_{lang_p}) \times z_s \times z_o \,.
\label{eq:alpha_product}
\end{align}

In the experiments of this section, the UVTransE hyperparameters are $C = 0.1$ and $\alpha = 0.5$. 

\smallskip
\noindent{\bf Evaluation metrics.}
To evaluate on the VG-IMP subset, we follow a methodology consistent with \cite{xu2017scenegraph} and report performance for the following three settings.
\begin{enumerate}
    \item {\bf Predicate Classification (PredCls)}: Given ground truth boxes and their corresponding objects, predict the predicate between object pairs. This is the same as predicate detection of Section \ref{subsec:vrd}.
    \item {\bf Phrase Classification (PhrCls)}: Given ground truth boxes, recognize the objects and their relations.
    \item {\bf Scene Graph Generation (SGGen)}: Predict objects, boxes (IoU $\geq 0.5$) and the relations between object pairs directly from an image. This is equivalent to relationship detection in Section \ref{subsec:vrd}.
\end{enumerate}
On the VG-VTransE subset, we follow \cite{Zhang_2017_CVPR} and report the performance for phrase and relationship detection, defined as in section \ref{subsec:vrd}.

For both subsets, we use Recall@50 and Recall@100 to evaluate how many labelled relationships are hit in the top 50 or 100 predictions. We follow related works in enforcing that for a given subject and object bounding box, the system must not output multiple predicate labels, which is the same as setting $k = 1$ in the VRD dataset~\cite{lu2016visual}.

\smallskip\noindent{\bf Comparison with state of the art.}
Table  \ref{table:vg_model_comparison} summarizes different state-of-the-art methods on the VG-IMP subset.
{\bf IMP}~\cite{xu2017scenegraph} uses standard RNNs and learns to iteratively improve its predictions via message passing between predicates. {\bf MSDN}~\cite{li2017msdn} jointly refines the features for different tasks by passing messages along a dynamically constructed graph. {\bf Neural Motifs}~\cite{zellers2018scenegraphs} proposes a Stacked Motif Network to capture higher-order motifs in scene graphs. {\bf Graph R-CNN}~\cite{jwyang_graph_rcnn} utilizes attentional graph convolutional networks to learn to modulate information flow through unlikely edges in the scene graph.

\begin{table*}[tbp!]
 \begin{center}
  \resizebox{\textwidth}{!}{
  \begin{tabular}{ p{7.0cm} l c c c c c}
   \hline
   \multicolumn{1}{p{7.0cm}}{} & Detector (pre-training) & mAP & ROI feature & Spatial feature & Language feature & Joint reasoning\\\hline
   Iterative Message Passing ({\bf IMP}, 2017) \cite{xu2017scenegraph} & VGG (COCO) &  & Pool   & - & - & \checkmark \\
   Multi-level Scene Description Network ({\bf MSDN}, 2017) \cite{li2017msdn} & VGG (ImageNet) &  & Pool & - & - & \checkmark \\
   {\bf Neural Motifs} (2018) \cite{zellers2018scenegraphs} & VGG (ImageNet) & 20.0 & Align & \checkmark & \checkmark & \checkmark \\
   {\bf Graph R-CNN} (2018) \cite{jwyang_graph_rcnn} & VGG (ImageNet) & 20.4  & Align & \checkmark & - & \checkmark  \\
   Relationship Detection Network  ({\bf RelDN}, 2019)~\cite{zhang2019vrd} & VGG (COCO) & 25.5 & Align & \checkmark & \checkmark & -  \\
   Large Scale Visual Relationship ({\bf LS-VRD}, 2019)~\cite{zhang2018large} & VGG (COCO) & - & - & - & \checkmark & - \\
   \hline
   {\bf UVTransE [VGG+V+L]} & VGG (ImageNet) & 19.2 & Align & \checkmark & \checkmark & - \\
   {\bf UVTransE [ResNet+V+L]} & ResNet (ImageNet) & 23.8 & Align & \checkmark & \checkmark & -  \\
   \hline
  \end{tabular}
  }
 \end{center}
 \caption{
 Summary of state-of-the-art methods on the VG-IMP dataset. See caption of Table \ref{table:vrd_model_comparison} for explanation of the columns.} 
 \label{table:vg_model_comparison}
\end{table*}

\begin{table*}[tbp!]
 \begin{center}
  \begin{tabular}{ p{7.0cm} c c c c c c}
   \hline
   \multicolumn{1}{p{7.0cm}}{} & 
   \multicolumn{2}{c}{PredCls} & \multicolumn{2}{c}{PhrCls} & \multicolumn{2}{c}{SGGen}\\
   & R@50& R@100 &R@50& R@100 &R@50& R@100 \\
   \hline
   {\bf IMP} \cite{xu2017scenegraph} & 44.8 & 53.1 & 21.7 & 24.4 & 3.4 & 4.2\\
   {\bf MSDN} \cite{li2017msdn} & 63.1 & 66.4 & 19.3 & 21.8 & 7.7 & 10.5\\
   {\bf Neural Motifs} \cite{zellers2018scenegraphs} & 65.2 & 67.1 & 35.8 & 36.5 & 27.2 & 30.3\\
   {\bf Graph R-CNN} \cite{jwyang_graph_rcnn} & 54.2 & 59.1 & 29.6 & 31.6 & 11.4 & 13.7\\
   {\bf RelDN}~\cite{zhang2019vrd} & {\bf 68.4} & {\bf 68.4} & {\bf 36.8} & {\bf 36.8} & 28.3 &  32.7\\
   {\bf LS-VRD}~\cite{zhang2018large} & {\bf 68.4} & {\bf 68.4} & \underline{36.7} & \underline{36.7} & 27.9 &  32.5\\
   \hline
   {\bf UVTransE [VGG+V]} & 59.7 & 63.3 & 30.7 & 31.9 & 25.2 & 28.3\\
   {\bf UVTransE [VGG+V+L]} & 61.2 & 64.3 & 30.9 & 32.2 & 25.3 & 28.5\\
   {\bf UVTransE [ResNet+V]} & 64.4 & 66.5 & 35.0 & 36.1 & \underline{29.9} & \underline{33.2}\\
   {\bf UVTransE [ResNet+V+L]} & \underline{65.3} & \underline{67.3} & 35.9 & 36.6 & {\bf 30.1} & {\bf 33.6} \\\hline
  \end{tabular}
 \end{center}
 \caption{Full test set performance on the VG-IMP dataset. {\bf Bold} indicates highest numbers, \underline{underline} indicates second-highest.} 
 \label{table:vg_imp}
\end{table*}

Comparative evaluation results on VG-IMP are shown in Table \ref{table:vg_imp}.  
We can see that our model with the ResNet detector ({\bf UVTransE (ResNet+V+L)}) outperforms all methods except the very recent RelDN, whose detector is even more accurate than ours, and LS-VRD. In particular, we get better performance than several methods that include message passing or graph CNNs to jointly reason about multiple relationships.
We also observe that our language module does not enjoy as significant a gain as in Table \ref{table:vrd_seen}. This is likely due to the fact that there are far more relations in the Visual Genome dataset than in VRD, so the training data for the language model is sparser. 

Table \ref{table:vg_vtranse} reports results on the VG-VTransE subset, which has an even larger number of object classes and relationships than VG-IMP. Here, as in Section \ref{subsec:vrd}, we can once again observe significant improvements over VTransE.

\begin{table}[tp!]
 \begin{center}
  \resizebox{\columnwidth}{!}{
  \begin{tabular}{ l c c c c}
   \hline
   \multicolumn{1}{l}{} & \multicolumn{2}{c}{Phr. Det.} & \multicolumn{2}{c}{Rel. Det.}\\
   &R@50& R@100 &R@50& R@100 \\
   \hline
   {\bf VTransE}~\cite{Zhang_2017_CVPR} & 9.46 & 10.45 & 5.52 & 6.04 \\\hline
   {\bf UVTransE [V]} & 15.47 & 19.70 &  8.52 &  10.59 \\
   {\bf UVTransE [V+L]} & {\bf 17.53} & {\bf 21.92} & {\bf 9.55} & {\bf 11.74} \\\hline
  \end{tabular}
  }
 \end{center}
 \caption{Full test set performance on the VG-VTransE dataset. {\bf Bold} indicates highest numbers.} 
 \label{table:vg_vtranse}
\end{table}

\smallskip 

\begin{figure*}[t!]
\centering
    \includegraphics[trim={0 0cm 0 0cm}, clip, width=0.75\textwidth, height=0.55\textheight ]{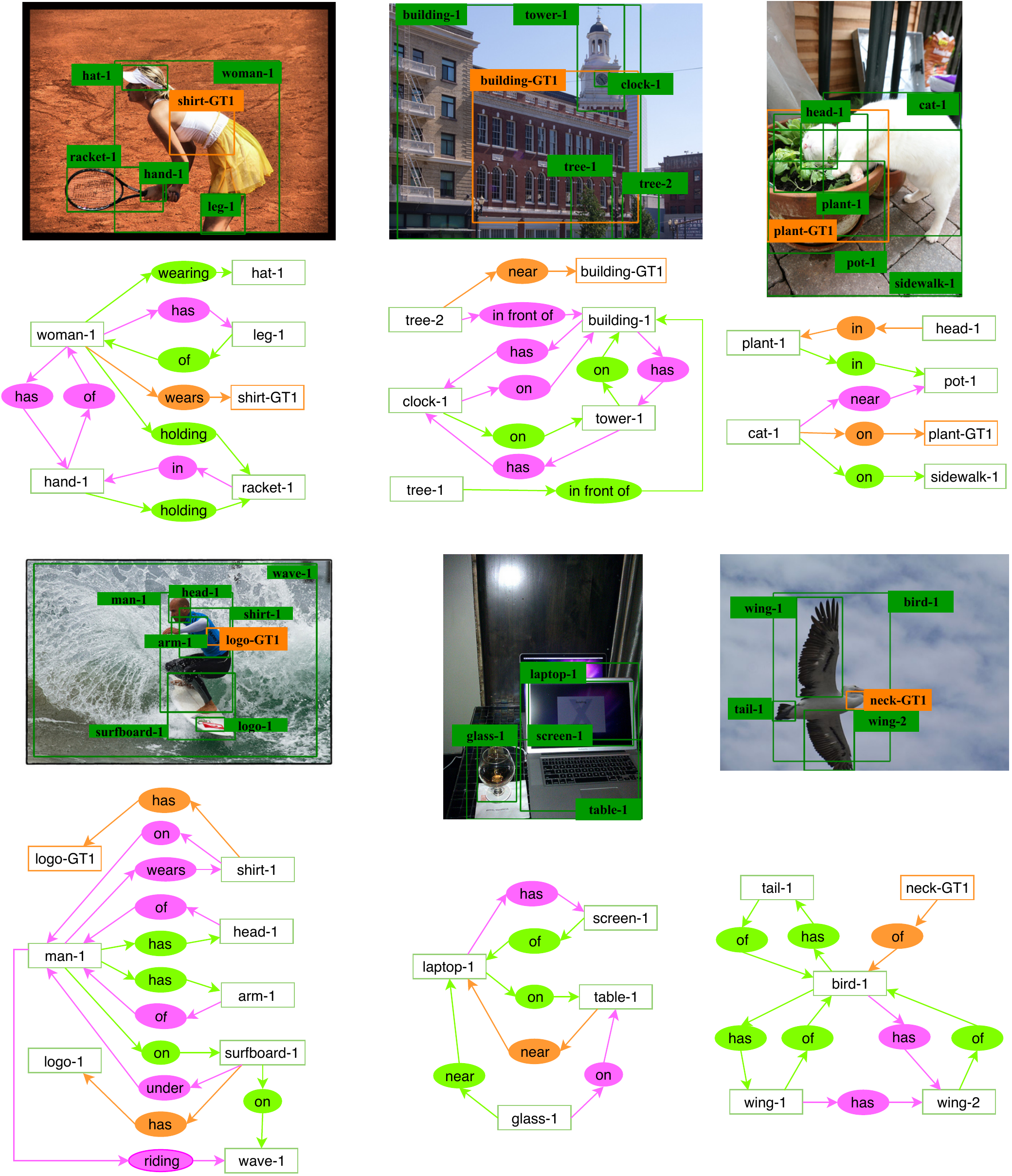}
   \caption{Example scene graphs generated on VG-IMP images. In the images, green boxes are objects detected with $IoU \geq 0.5$, while orange boxes are ground truth objects that are not detected by our pipeline. In the scene graphs,  green ellipses are true positive relations recognized by our model at Recall@20, orange ellipses are false negatives, and magenta ellipses are false positives (sometimes due to missing ground truth).}
\label{fig:vg_res}
\end{figure*}

\noindent {\bf Qualitative results of scene graph generation.}
In Figure \ref{fig:vg_res}, we show some example outputs of scene graph generation using {\bf UVTransE[V+L]} on Visual Genome. Through careful inspection, we can see that UVTransE generally fails in two cases: either the object detector cannot find the objects present in the ground truth, which are highlighted with orange boxes, or the spatial configuration makes it hard to predict the predicate. For instance, in the image with the pelican (bottom right), there is a predicted false positive: (\emph{wing-1}, \emph{has}, \emph{wing-2}).
In addition, many seemingly correct relations are marked as false positives due to incomplete ground truth. For example, (\emph{racket-1}, \emph{in}, \emph{hand-1}) is a plausible relation in the top left image of Figure \ref{fig:vg_res}; however, it does not exist in the annotations. 

\begin{figure*}[t!]
\centering
    \includegraphics[trim={0 0cm 0 0cm}, clip,width=0.75\textwidth, height=0.5\textheight]{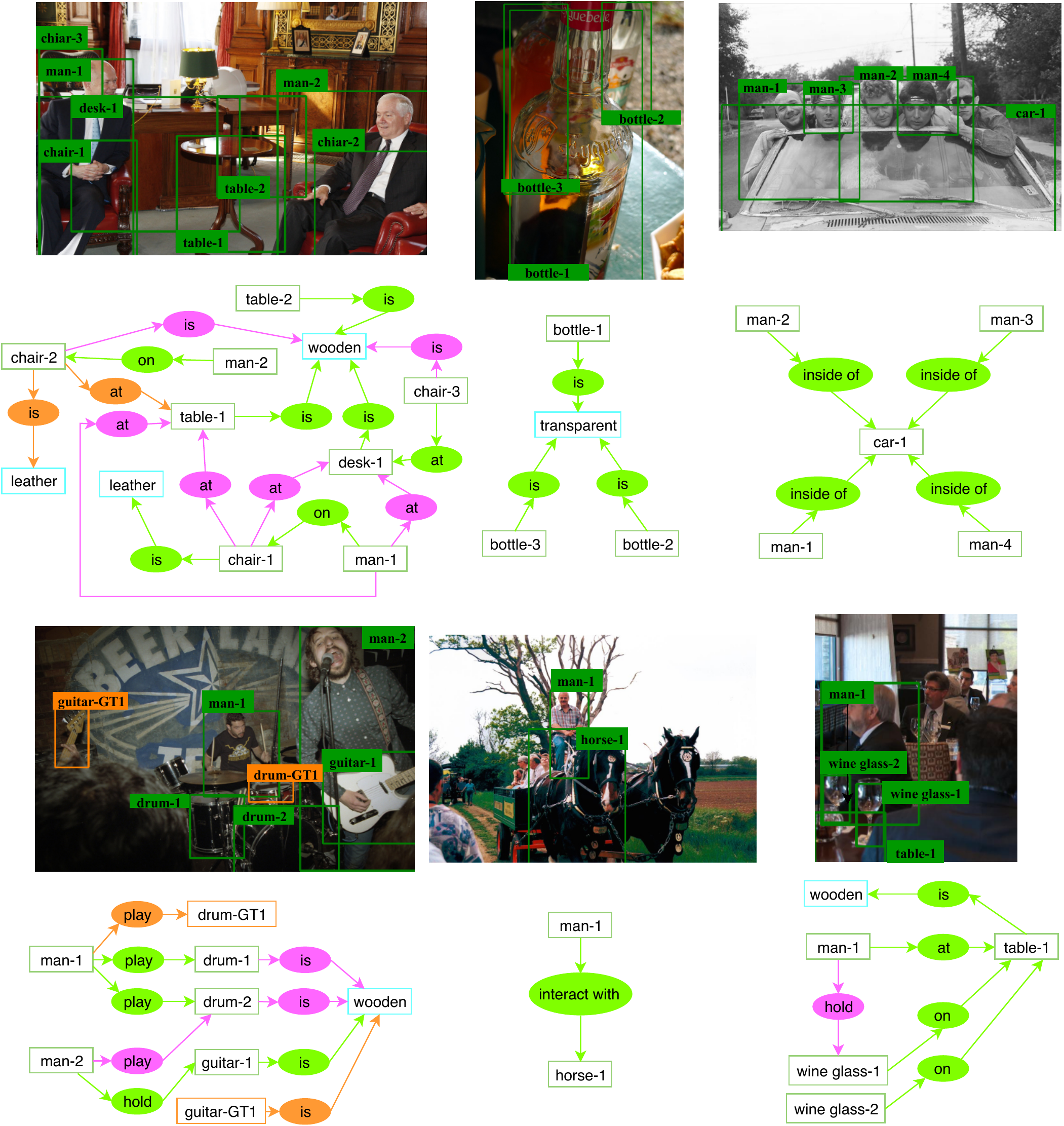}
   \caption{Example scene graphs generated on Open Images. In the images, green boxes are objects detected with $IoU \geq 0.5$, while orange ones are ground truth objects that are not detected. In the scene graphs, attributes are represented with cyan boxes. Green ellipses are true positive relations recognized by our model at Recall@20,  orange ellipses are false negatives, and magenta ellipses are false positives.}
\label{fig:openimage_res}
\end{figure*}

\subsection{{\bf Results on the Open Images Dataset}} \label{subsec:openimages}
\noindent{\bf Dataset.} Our final set of experiments is on the Open Images dataset~\cite{OpenImages}, which is even larger than Visual Genome: 94,747 training and 5,775 validation images according to the recommended split. On the other hand, the number of object classes and predicates is smaller, only 57 and 10, respectively. Among the 10 predicates, there is one special predicate, \emph{is}, which is used to describe visual attributes, e.g., (\emph{table}, \emph{is}, \emph{wooden}). Therefore, in addition to relation prediction, we also have to adapt our method to perform attribute prediction. 

\smallskip\noindent{\bf Implementation details.}
We use Faster R-CNN with ResNet-101 backbone as the object detector and region feature extractor. We initialize the network with weights pre-trained on COCO~\cite{lin2014microsoft} and fine-tune on Open Images to achieve an mAP of 51\% on our validation split. Similar to the Visual Genome setup described in section \ref{sec:vg_setup}, we freeze the detector and only train our UVTransE module along with the language module with $C = 0.1$ and $\alpha = 0.5$. During training, 25\% of triplets in each batch are positive. In test time, we select the top 50 candidate proposals from Faster R-CNN for mining relationships, and the final triplet scores are calculated with  Eq. (\ref{eq:vis_only_score_product}) and Eq. (\ref{eq:alpha_product}) for UVTransE [V] and UVTransE [V+L] predictions, respectively.

In order to tackle the predicate \emph{is}, we use the same object proposals generated by Faster R-CNN and train an additional classifier on each proposal to output the probability for each attribute. The attribute score is calculated with
\begin{equation}
z_{\left(s, \emph{is}, a\right)} = z_s \times z_a ,
\label{eq:attr_score}
\end{equation}
where $a$ is the attribute and $z_a$ is the output probability from the attribute classifier. 

\smallskip\noindent{\bf Evaluation metrics.}
In the Open Images Challenge, results are evaluated based on Recall@50 of relationship detection (R@N$_{rel}$), mAP of relationship detection (mAP$_{rel}$), and mAP of phrase detection (mAP$_{phr}$).  
The final score is calculated with $0.2 \times R@N_{rel} + 0.4 \times mAP_{rel} + 0.4 \times mAP_{phr}$. The mAP$_{rel}$ takes the mean of AP for each predicate, where true positive is defined as having correct object boxes ($IoU \geq 0.5$), classes, and predicates. The mAP$_{phr}$ is similar to mAP$_{rel}$, but applied to the union of subject and object boxes instead of individual boxes.

\smallskip\noindent{\bf Comparison with state of the art.} We compare our results with other models from the official kaggle competition. There are 99,999 test images, and the official test set is split into public and private sets, which contain 30\% and 70\% of test data, respectively. We present results for both splits in Table \ref{table:openimage}. We also have an additional column, named ``Full", for overall performance, which is calculated by $0.3 \times$ {\em  public\_score} $+ 0.7 \times $ {\em private\_score}. 
As shown in Table \ref{table:openimage}, we surpass most teams except for RelDN, who once again use a better object detector (Faster R-CNN with ResNeXt-101-FPN).
Notice also the large gap between {\bf UVTransE [V+L]} and the second place ({\bf kyle}) considering the low absolute scores and the large amount of test images. 

\begin{table}[t!]
 \begin{center}
  \begin{tabular}{ p{3.0cm} c c c}
   \hline
   \multicolumn{1}{p{3.0cm}}{} &
   \multicolumn{1}{c}{Public} &
   \multicolumn{1}{c}{Private} &
   \multicolumn{1}{c}{Full}\\
   \hline
   {\bf tito} & 0.256 & 0.237 & 0.243\\
   {\bf kyle} & 0.280 & 0.235 & 0.249\\
   {\bf RelDN}~\cite{zhang2019vrd} & \underline{0.320} & {\bf 0.332} & {\bf 0.328}\\\hline
   {\bf UVTransE [V]} & 0.285 & 0.246 & 0.258\\
   {\bf UVTransE [V+L]} & {\bf 0.321} & \underline{0.273} & \underline{0.287}\\\hline
  \end{tabular}
 \end{center}
 \caption{Results on Open Images Challenge for the top three teams on the public leaderboard vs. our methods. These values are evaluated based on the official mAP$_{rel}$, mAP$_{phr}$, and Recall@50 for relationship detection. Public and private correspond to 30\% and 70\% of test data respectively. Full is $0.3 \times $ {\em public} $+ 0.7 \times$ {\em private}. {\bf Bold} indicates highest numbers. \underline{Underline} indicates second highest.} 
 \label{table:openimage}
\end{table}

\smallskip
\noindent{\bf Qualitative results of scene graph generation.} 
Figure \ref{fig:openimage_res} presents examples of generated scene graphs on Open Images. Our model is able to cover different kinds of relations, including positional predicates such as \emph{on}, attributive predicates such as \emph{is}, and interactive predicates such as \emph{play}. Similar to the results on Visual Genome, UVTransE has a hard time when the spatial configuration is challenging. Take the top left image, which contains two people sitting on chairs as an example. We can see that our model outputs (\emph{man-1}, \emph{at}, \emph{table-1}), whose spatial structure is quite similar to other relationships that involve \emph{at}, such as (\emph{chair}, \emph{at}, \emph{desk}).

\section{Conclusion}

In this paper, we introduced the UVTransE framework for visual relationship detection, which extends the VTransE framework~\cite{Zhang_2017_CVPR} by adding a union box feature to the subject and object box features for learning the embedding of the predicate. While our original motivation was primarily to improve zero-shot performance of VTransE, extensive experiments have demonstrated that our UVTransE model achieves state-of-the-art results in multiple challenging scenarios, from small-scale to large-scale, on both the full test set and zero-shot settings. The latter is a significant contribution, since some other state-of-the-art methods, like RelDN~\cite{zhang2019vrd} achieve high accuracy on common relationships at the cost of low zero-shot performance. We obtain consistent improvements over prior work while keeping the formulation straightforward. 
The simplicity of our model combined with its versatility and high performance thus makes it a good practical choice for advanced visual reasoning tasks such as scene graph generation.



%





\ifCLASSOPTIONcaptionsoff
  \newpage
\fi



\bibliographystyle{IEEEtran}
\bibliography{egbib}

\begin{thebibliography}{10}
\providecommand{\url}[1]{#1}
\csname url@samestyle\endcsname
\providecommand{\newblock}{\relax}
\providecommand{\bibinfo}[2]{#2}
\providecommand{\BIBentrySTDinterwordspacing}{\spaceskip=0pt\relax}
\providecommand{\BIBentryALTinterwordstretchfactor}{4}
\providecommand{\BIBentryALTinterwordspacing}{\spaceskip=\fontdimen2\font plus
\BIBentryALTinterwordstretchfactor\fontdimen3\font minus
  \fontdimen4\font\relax}
\providecommand{\BIBforeignlanguage}[2]{{%
\expandafter\ifx\csname l@#1\endcsname\relax
\typeout{** WARNING: IEEEtran.bst: No hyphenation pattern has been}%
\typeout{** loaded for the language `#1'. Using the pattern for}%
\typeout{** the default language instead.}%
\else
\language=\csname l@#1\endcsname
\fi
#2}}
\providecommand{\BIBdecl}{\relax}
\BIBdecl

\bibitem{Zhang_2017_CVPR}
H.~Zhang, Z.~Kyaw, S.-F. Chang, and T.-S. Chua, ``Visual translation embedding
  network for visual relation detection,'' in \emph{CVPR}, 2017.

\bibitem{girshick2014rich}
R.~Girshick, J.~Donahue, T.~Darrell, and J.~Malik, ``Rich feature hierarchies
  for accurate object detection and semantic segmentation,'' in \emph{CVPR},
  2014.

\bibitem{girshick15fastrcnn}
R.~Girshick, ``Fast {R-CNN},'' in \emph{ICCV}, 2015.

\bibitem{he2017mask}
K.~He, G.~Gkioxari, P.~Doll{\'a}r, and R.~Girshick, ``Mask r-cnn,'' in
  \emph{ICCV}, 2017.

\bibitem{ren2015faster}
S.~Ren, K.~He, R.~Girshick, and J.~Sun, ``Faster {R-CNN}: Towards real-time
  object detection with region proposal networks,'' in \emph{NIPS}, 2015.

\bibitem{redmon2017yolo9000}
J.~Redmon and A.~Farhadi, ``Yolo9000: better, faster, stronger,'' in
  \emph{CVPR}, 2016.

\bibitem{lu2018neural}
J.~Lu, J.~Yang, D.~Batra, and D.~Parikh, ``Neural baby talk,'' in \emph{CVPR},
  2018.

\bibitem{sg2015}
J.~Johnson, R.~Krishna, M.~Stark, J.~Li, M.~Bernstein, and L.~Fei-Fei, ``Image
  retrieval using scene graphs,'' in \emph{CVPR}, 2015.

\bibitem{PrabhuICCV2015}
N.~Prabhu and R.~V. Babu, ``Attribute-graph: A graph based approach to image
  ranking,'' in \emph{ICCV}, 2015.

\bibitem{nmn2016}
J.~Andreas, M.~Rohrbach, T.~Darrell, and D.~Klein, ``Deep compositional
  question answering with neural module networks,'' in \emph{CVPR}, 2016.

\bibitem{lu2016visual}
C.~Lu, R.~Krishna, M.~Bernstein, and L.~Fei-Fei, ``Visual relationship
  detection with language priors,'' in \emph{ECCV}, 2016.

\bibitem{bordes2013}
A.~Bordes, N.~Usunier, A.~Garcia-Duran, J.~Weston, and O.~Yakhnenko,
  ``Translating embeddings for modeling multi-relational data,'' in
  \emph{NIPS}, 2013.

\bibitem{plummerPLCLC2017}
B.~A. Plummer, A.~Mallya, C.~M. Cervantes, J.~Hockenmaier, and S.~Lazebnik,
  ``Phrase localization and visual relationship detection with comprehensive
  image-language cues,'' in \emph{ICCV}, 2017.

\bibitem{NIPS2013_5021}
T.~Mikolov, I.~Sutskever, K.~Chen, G.~S. Corrado, and J.~Dean, ``Distributed
  representations of words and phrases and their compositionality,'' in
  \emph{NIPS}, 2013.

\bibitem{pennington2014glove}
J.~Pennington, R.~Socher, and C.~D. Manning, ``Glove: Global vectors for word
  representation.'' in \emph{EMNLP}, 2014.

\bibitem{Peyre17}
J.~Peyre, I.~Laptev, C.~Schmid, and J.~Sivic, ``Weakly-supervised learning of
  visual relations,'' in \emph{ICCV}, 2017.

\bibitem{Krishna:2017:VGC:3088990.3089101}
R.~Krishna, Y.~Zhu, O.~Groth, J.~Johnson, K.~Hata, J.~Kravitz, S.~Chen,
  Y.~Kalantidis, L.-J. Li, D.~A. Shamma, M.~S. Bernstein, and L.~Fei-Fei,
  ``Visual genome: Connecting language and vision using crowdsourced dense
  image annotations,'' \emph{IJCV}, 2017.

\bibitem{OpenImages}
A.~Kuznetsova, H.~Rom, N.~Alldrin, J.~Uijlings, I.~Krasin, J.~Pont-Tuset,
  S.~Kamali, S.~Popov, M.~Malloci, T.~Duerig, and V.~Ferrari, ``The open images
  dataset v4: Unified image classification, object detection, and visual
  relationship detection at scale,'' \emph{arXiv:1811.00982}, 2018.

\bibitem{zellers2018scenegraphs}
R.~Zellers, M.~Yatskar, S.~Thomson, and Y.~Choi, ``Neural motifs: Scene graph
  parsing with global context,'' in \emph{{CVPR}}, 2018.

\bibitem{Galleguillos08}
C.~Galleguillos, A.~Rabinovich, and S.~Belongie, ``Object categorization using
  co-occurrence, location and appearance,'' in \emph{CVPR}, 2008.

\bibitem{Gould2008}
S.~Gould, J.~Rodgers, D.~Cohen, G.~Elidan, and D.~Koller, ``Multi-class
  segmentation with relative location prior,'' \emph{IJCV}, 2008.

\bibitem{gkioxari2015rstarcnn}
G.~Gkioxari, R.~Girshick, and J.~Malik, ``Contextual action recognition with
  r\*cnn,'' in \emph{ICCV}, 2015.

\bibitem{Guadarrama:2013}
S.~Guadarrama, N.~Krishnamoorthy, G.~Malkarnenkar, S.~Venugopalan, R.~Mooney,
  T.~Darrell, and K.~Saenko, ``Youtube2text: Recognizing and describing
  arbitrary activities using semantic hierarchies and zero-shot recognition,''
  in \emph{ICCV}, 2013.

\bibitem{mallya2016learning}
A.~Mallya and S.~Lazebnik, ``Learning models for actions and person-object
  interactions with transfer to question answering,'' in \emph{ECCV}, 2016.

\bibitem{maji2011action}
S.~Maji, L.~Bourdev, and J.~Malik, ``Action recognition from a distributed
  representation of pose and appearance,'' in \emph{CVPR}, 2011.

\bibitem{yao2011human}
B.~Yao, X.~Jiang, A.~Khosla, A.~L. Lin, L.~Guibas, and L.~Fei-Fei, ``Human
  action recognition by learning bases of action attributes and parts,'' in
  \emph{ICCV}, 2011.

\bibitem{YaoF10}
B.~Yao and F.-F. Li, ``Grouplet: A structured image representation for
  recognizing human and object interactions.'' in \emph{CVPR}, 2010.

\bibitem{choi_cvpr10}
M.~J. Choi, J.~J. Lim, A.~Torralba, and A.~S. Willsky, ``Exploiting
  hierarchical context on a large database of object categories,'' in
  \emph{CVPR}, 2010.

\bibitem{Papazoglou:2016:DOA:2997676.2998110}
A.~Papazoglou, L.~D. Pero, and V.~Ferrari, ``Discovering object aspects from
  video,'' \emph{Image and Vision Computing}, 2016.

\bibitem{SalakhutdinovTT11}
R.~Salakhutdinov, A.~Torralba, and J.~B. Tenenbaum, ``Learning to share visual
  appearance for multiclass object detection,'' in \emph{{CVPR}}, 2011.

\bibitem{MensinkGS14}
T.~Mensink, E.~Gavves, and C.~G.~M. Snoek, ``{COSTA:} co-occurrence statistics
  for zero-shot classification,'' in \emph{{CVPR}}, 2014.

\bibitem{fukui16emnlp}
A.~Fukui, D.~H. Park, D.~Yang, A.~Rohrbach, T.~Darrell, and M.~Rohrbach,
  ``Multimodal compact bilinear pooling for visual question answering and
  visual grounding,'' in \emph{{EMNLP}}, 2016.

\bibitem{Plummer:2017:_flickr}
B.~A. Plummer, L.~Wang, C.~M. Cervantes, J.~C. Caicedo, J.~Hockenmaier, and
  S.~Lazebnik, ``Flickr30k entities: Collecting region-to-phrase
  correspondences for richer image-to-sentence models,'' \emph{IJCV}, 2017.

\bibitem{dai2017detecting}
B.~Dai, Y.~Zhang, and D.~Lin, ``Detecting visual relationships with deep
  relational networks,'' in \emph{CVPR}, 2017.

\bibitem{vrl17}
X.~Liang, L.~Lee, and E.~P. Xing, ``Deep variation-structured reinforcement
  learning for visual relationship and attribute detection,'' in \emph{CVPR},
  2017.

\bibitem{zhang2018large}
J.~Zhang, Y.~Kalantidis, M.~Rohrbach, M.~Paluri, A.~Elgammal, and M.~Elhoseiny,
  ``Large-scale visual relationship understanding,'' in \emph{AAAI}, 2019.

\bibitem{zhang2019vrd}
J.~Zhang, K.~J. Shih, A.~Elgammal, A.~Tao, and B.~Catanzaro, ``Graphical
  contrastive losses for scene graph generation,'' in \emph{CVPR}, 2019.

\bibitem{zhuang2017towards}
B.~Zhuang, L.~Liu, C.~Shen, and I.~Reid, ``Towards context-aware interaction
  recognition for visual relationship detection,'' in \emph{ICCV}, 2017.

\bibitem{yu2017_vrd_knowledge_distillation}
R.~Yu, A.~Li, V.~I. Morariu, and L.~S. Davis, ``Visual relationship detection
  with internal and external linguistic knowledge distillation.'' in
  \emph{ICCV}, 2017.

\bibitem{xu2017scenegraph}
D.~Xu, Y.~Zhu, C.~Choy, and L.~Fei-Fei, ``Scene graph generation by iterative
  message passing,'' in \emph{CVPR}, 2017.

\bibitem{li2017msdn}
Y.~Li, W.~Ouyang, B.~Zhou, K.~Wang, and X.~Wang, ``Scene graph generation from
  objects, phrases and region captions,'' in \emph{{ICCV}}, 2017.

\bibitem{jwyang_graph_rcnn}
J.~Yang, J.~Lu, S.~Lee, D.~Batra, and D.~Parikh, ``Graph {R-CNN} for scene
  graph generation,'' in \emph{{ECCV}}, 2018.

\bibitem{Schuster:1997:BRN:2198065.2205129}
M.~Schuster and K.~Paliwal, ``Bidirectional recurrent neural networks,''
  \emph{TSP}, 1997.

\bibitem{simonyan14VGG}
K.~Simonyan and A.~Zisserman, ``Very deep convolutional networks for
  large-scale image recognition,'' \emph{CoRR}, vol. abs/1409.1556, 2014.

\bibitem{he2016deep}
K.~He, X.~Zhang, S.~Ren, and J.~Sun, ``Deep residual learning for image
  recognition,'' in \emph{CVPR}, 2016.

\bibitem{liang2018Visual}
K.~Liang, Y.~Guo, H.~Chang, and X.~Chen, ``Visual relationship detection with
  deep structural ranking,'' in \emph{AAAI}, 2018.

\bibitem{gjyin_eccv2018}
G.~Yin, L.~Sheng, B.~Liu, N.~Yu, X.~Wang, J.~Shao, and C.~C. Loy, ``Zoom-net:
  Mining deep feature interactions for visual relationship recognition,'' in
  \emph{ECCV}, 2018.

\bibitem{lin2014microsoft}
T.-Y. Lin, M.~Maire, S.~Belongie, J.~Hays, P.~Perona, D.~Ramanan,
  P.~Doll{\'a}r, and C.~L. Zitnick, ``Microsoft coco: Common objects in
  context,'' in \emph{ECCV}, 2014.

\bibitem{densecap}
J.~Johnson, A.~Karpathy, and L.~Fei-Fei, ``Densecap: Fully convolutional
  localization networks for dense captioning,'' in \emph{CVPR}, 2016.

\end{thebibliography}
%



%







\end{document}